\documentclass[runningheads]{llncs}
\usepackage{graphicx}

\usepackage{tikz}
\usepackage{comment}
\usepackage{amsmath,amssymb} 
\usepackage{color}

\usepackage{epsfig}
\usepackage{caption}
\usepackage{subcaption}
\usepackage{float}
\usepackage[algo2e, ruled]{algorithm2e}
\SetKwInput{KwPara}{Parameter}

\usepackage{ulem}
\usepackage{hyperref}

\begin{document}
\pagestyle{headings}
\mainmatter
\def\ECCVSubNumber{36}  

\title{Weakly Supervised Minirhizotron Image Segmentation with MIL-CAM} 

%
\makeatletter
\newcommand*\fsize{\dimexpr\f@size pt\relax}
\makeatother
\renewcommand{\orcidID}[1]{\href{https://orcid.org/#1}{\includegraphics[height=1\fsize]{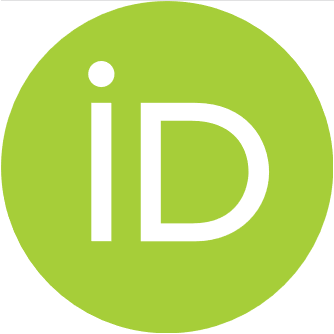}}}

\author{Guohao Yu\inst{1}\orcidID{0000-0002-6850-7241} \and
Alina Zare\inst{1}\orcidID{0000-0002-4847-7604} \and
Weihuang Xu\inst{1}\orcidID{0000--0001-8463-9319} \and
Roser Matamala \inst{2}\orcidID{0000-0001-5552-9807} \and
Joel Reyes-Cabrera \inst{3} \orcidID{0000-0002-3535-8665} \and
Felix B. Fritschi \inst{3} \orcidID{0000-0003-0825-6855} \and
Thomas E. Juenger \inst{4}\orcidID{0000-0001-9550-9288}
}
\authorrunning{G. Yu et al.}

%
\institute{Department of Electrical and Computer Engineering, University of Florida, Gainesville, FL, USA 32611
\email{azare@ece.ufl.edu}\and
Argonne National Laboratory, Lemont, IL, USA 60439 \and
Division of Plant Sciences, University of Missouri, Columbia, MO, USA 65211 \and
Department of Integrative Biology, University of Texas at Austin, Austin, TX, USA 78712}
\maketitle

\begin{abstract}
We present a multiple instance learning class activation map (MIL-CAM) approach for pixel-level minirhizotron image segmentation given weak image-level labels. Minirhizotrons are used to image plant roots \textit{in situ}.  Minirhizotron imagery is often composed of soil containing a few long and thin root objects of small diameter.  The roots prove to be challenging for existing semantic image segmentation methods to discriminate. In addition to learning from weak labels, our proposed MIL-CAM approach re-weights the root versus soil pixels during analysis for improved performance due to the heavy imbalance between soil and root pixels.  The proposed approach outperforms other attention map and multiple instance learning methods for localization of root objects in minirhizotron imagery.  \end{abstract}

\section{Introduction}

Minirhizotron (MR) imaging plays an important role in plant root studies. It is a widely-used non-destructive root sampling method used to monitor root systems over extended periods of time without repeatedly altering critical soil conditions or root processes \cite{johnson2001advancing,bates1937device,waddington1971observation,rewald2013minirhizotron}. Yet, a significant bottleneck which impacts the value of MR systems is the analysis time needed to process collected imagery.  Standard analysis approaches require manual root tracing and labeling of root characteristics.  Manually tracing roots collected with MR systems is very tedious, slow, and error prone.  Thus,  MR image analysis would greatly benefit from automated methods to segment and trace roots. There have been advancements made in this area \cite{wang2019segroot,xu2019overcoming,yu2019root}.  However, the effective automated methods still require a manually-labeled training set.  Although these approaches provide a reduction in effort needed over hand-tracing an entire collection of data, the generation of these training sets is still time consuming and labor intensive. In this paper, we propose a weakly supervised MR image segmentation method that relies only on image-level labels.  

By relying only on weak image level labels (e.g., this image does/does not contain roots), the time and labor needed to generate a training set is drastically reduced \cite{papandreou2015weakly,pinheiro2015image,zhou2016learning,roy2017combining,wei2018revisiting,tang2018normalized}.  It is also much easier and less error prone to identify when an image does or does not contain roots as opposed to correctly labeling every pixel in an image. However, current weakly-supervised methods used to infer pixel-levels labels do not perform as well as semantic image segmentation methods leveraging full annotation \cite{long2015fully,ronneberger2015u,lin2016efficient,chen2017deeplab,badrinarayanan2017segnet}. 
\begin{figure}[h] 
\begin{center}
\begin{subfigure}[t]{0.15\textwidth}
   \includegraphics[width=1\linewidth]{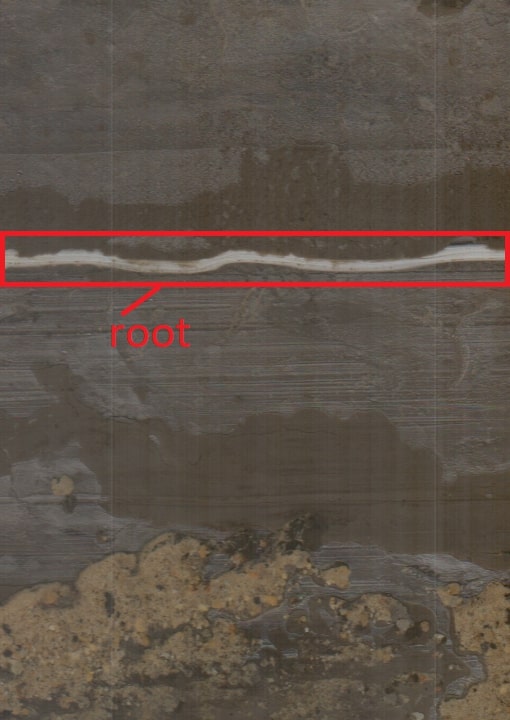}
   \caption{}
   \label{fig:1_1a}
\end{subfigure}
\begin{subfigure}[t]{0.15\textwidth}
   \includegraphics[width=1\linewidth]{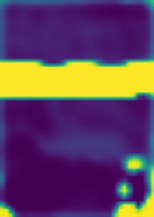}
   \caption{}
   \label{fig:1_1b}
\end{subfigure}
\begin{subfigure}[t]{0.15\textwidth}
   \includegraphics[width=1\linewidth]{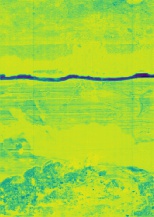}
   \caption{}
   \label{fig:1_1c}
\end{subfigure}
\begin{subfigure}[t]{0.15\textwidth}
   \includegraphics[width=1\linewidth]{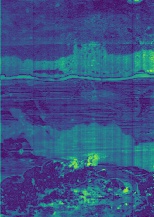}
  \caption{}
   \label{fig:1_1d}
\end{subfigure}
\begin{subfigure}[t]{0.15\textwidth}
   \includegraphics[width=1\linewidth]{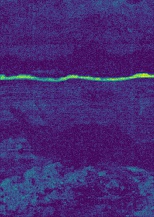}
   \caption{}
   \label{fig:1_1e}
\end{subfigure}
\begin{subfigure}[t]{0.15\textwidth}
   \includegraphics[width=1\linewidth]{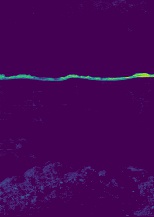}
  \caption{}
  \label{fig:1_1f}
\end{subfigure}

   \caption{Example attention maps from different methods of semantic segmentation of an MR image. (\subref{fig:1_1a}) Original MR Image. (\subref{fig:1_1b}) CAM Result (\subref{fig:1_1c}) Grad-CAM Result (\subref{fig:1_1d}) Grad-CAM++ Result (\subref{fig:1_1e})  SMOOTHGRAD Result (\subref{fig:1_1f}) Result of proposed method,\textbf{} MIL-CAM.}
\label{fig:1_1}
\end{center}
\end{figure}

Attention or class activation maps have been widely used to infer pixel-level labels from training data with weak image-level labels \cite{zhou2016learning,selvaraju2017grad,chattopadhay2018grad,smilkov2017smoothgrad}. However,  existing attention map approaches have been found to be inaccurate in identifying and delineating roots in MR imagery. For example, CAM (the class activation maps) approach \cite{zhou2016learning} overestimates the size of the roots as shown in Fig.\ref{fig:1_1b}. The Gradient-weighted Class Activation Mapping (Grad-CAM) \cite{selvaraju2017grad} approach incorrectly identifies the background soil as the root target as shown in Fig.\ref{fig:1_1c}. Grad-CAM++ \cite{chattopadhay2018grad}  and SMOOTHGRAD \cite{smilkov2017smoothgrad} shown in Fig.\ref{fig:1_1d} and Fig.\ref{fig:1_1e}, respectively, result in maps with poor contrast between roots and soil and, thus, many false alarms.

In this paper, we propose the multiple instance learning CAM (MIL-CAM) approach to address root segmentation in MR imagery given weak image-level labels.  MIL-CAM is outlined in Section \ref{sec3}. In Section \ref{sec4}, we compare MIL-CAM approach results to existing approaches with both weak- and full-annotation on an MR dataset collected from switchgrass.

\section{Related Work}
 \begin{figure}[htb] 
\begin{center}
\begin{subfigure}[t]{0.18\textwidth}
   \includegraphics[width=1\linewidth]{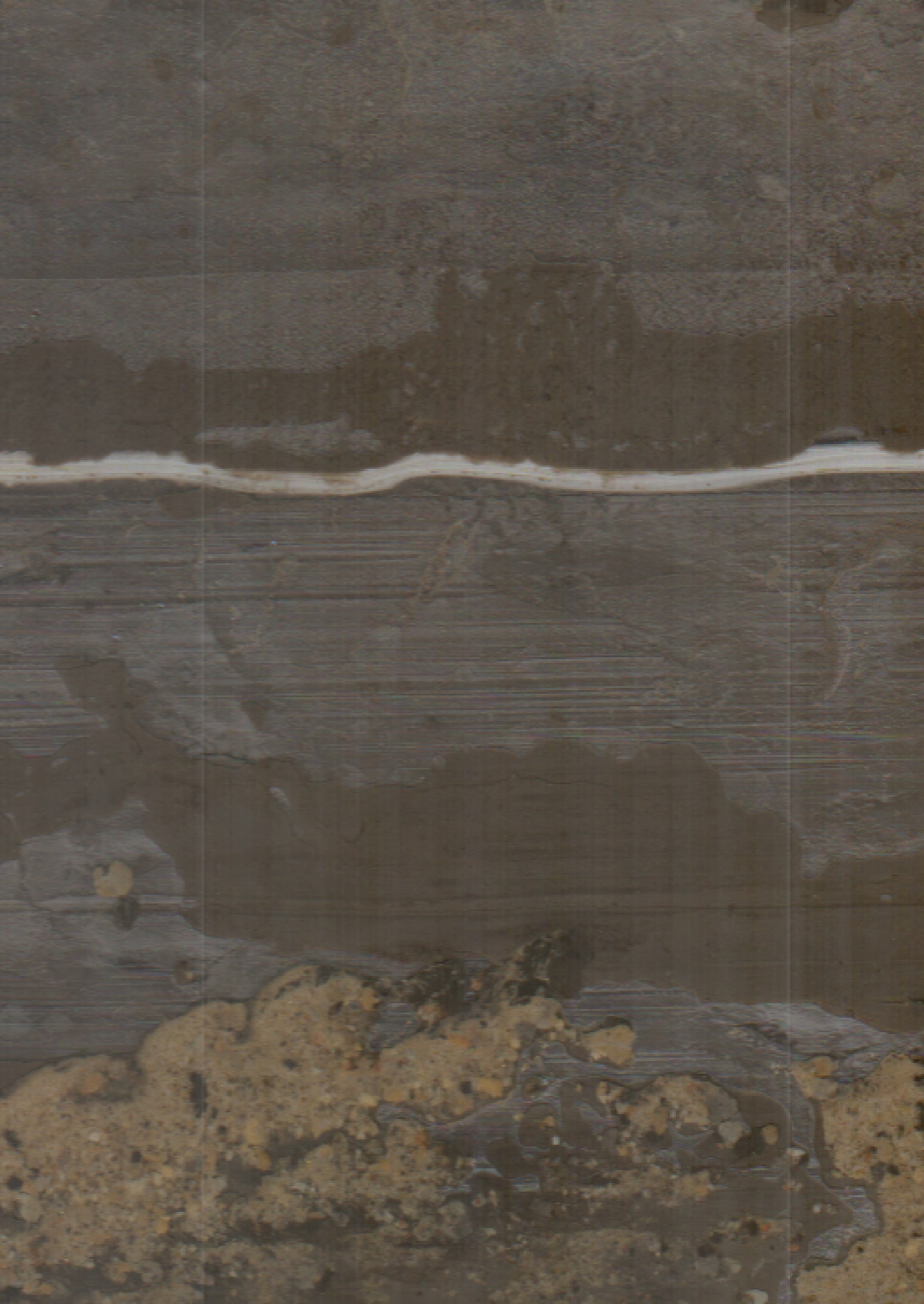}
   \caption{}
   \label{fig:8_1a}
\end{subfigure}
\begin{subfigure}[t]{0.18\textwidth}
   \includegraphics[width=1\linewidth]{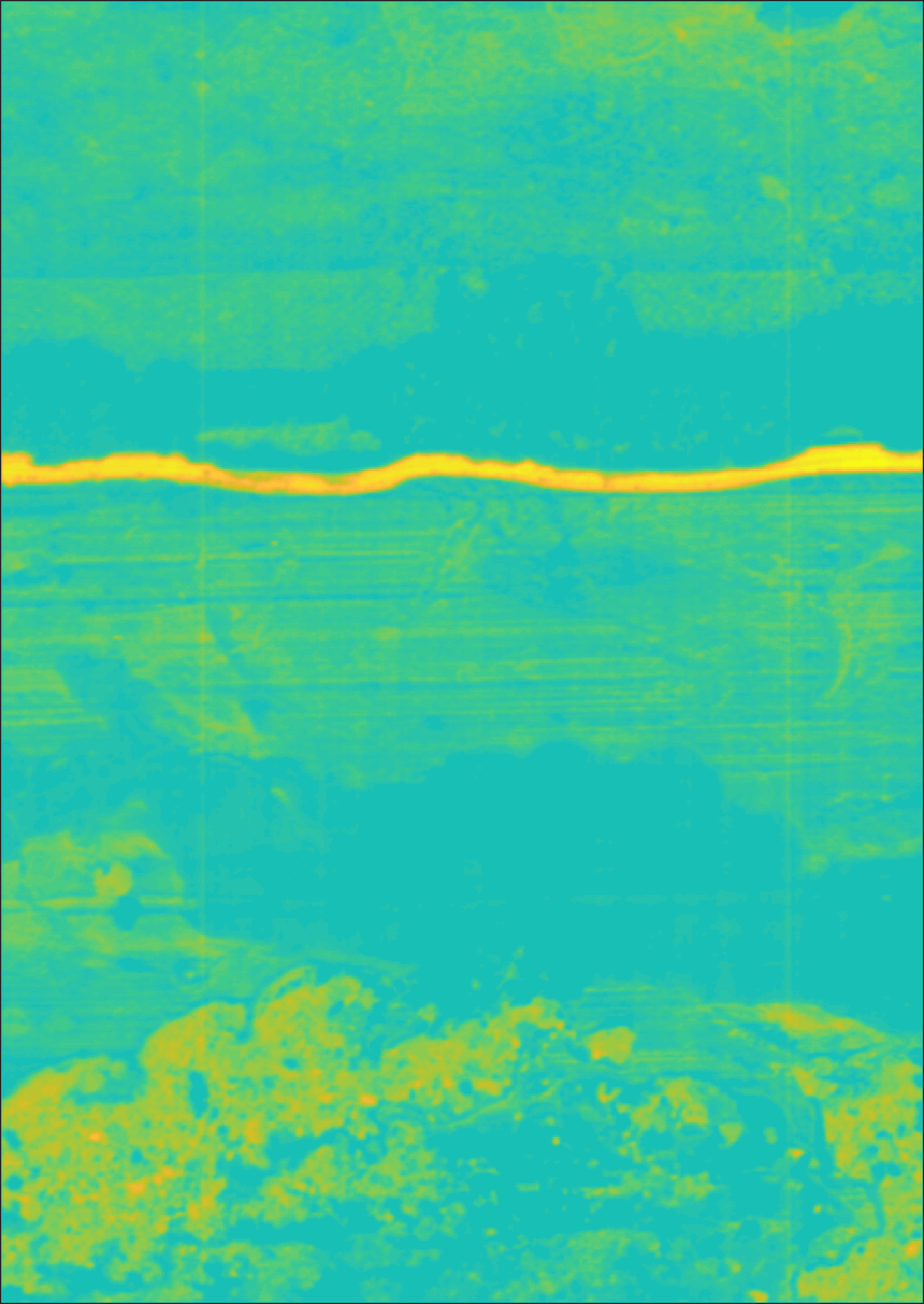}
   \caption{}
   \label{fig:8_1b}
\end{subfigure}
\begin{subfigure}[t]{0.18\textwidth}
   \includegraphics[width=1\linewidth]{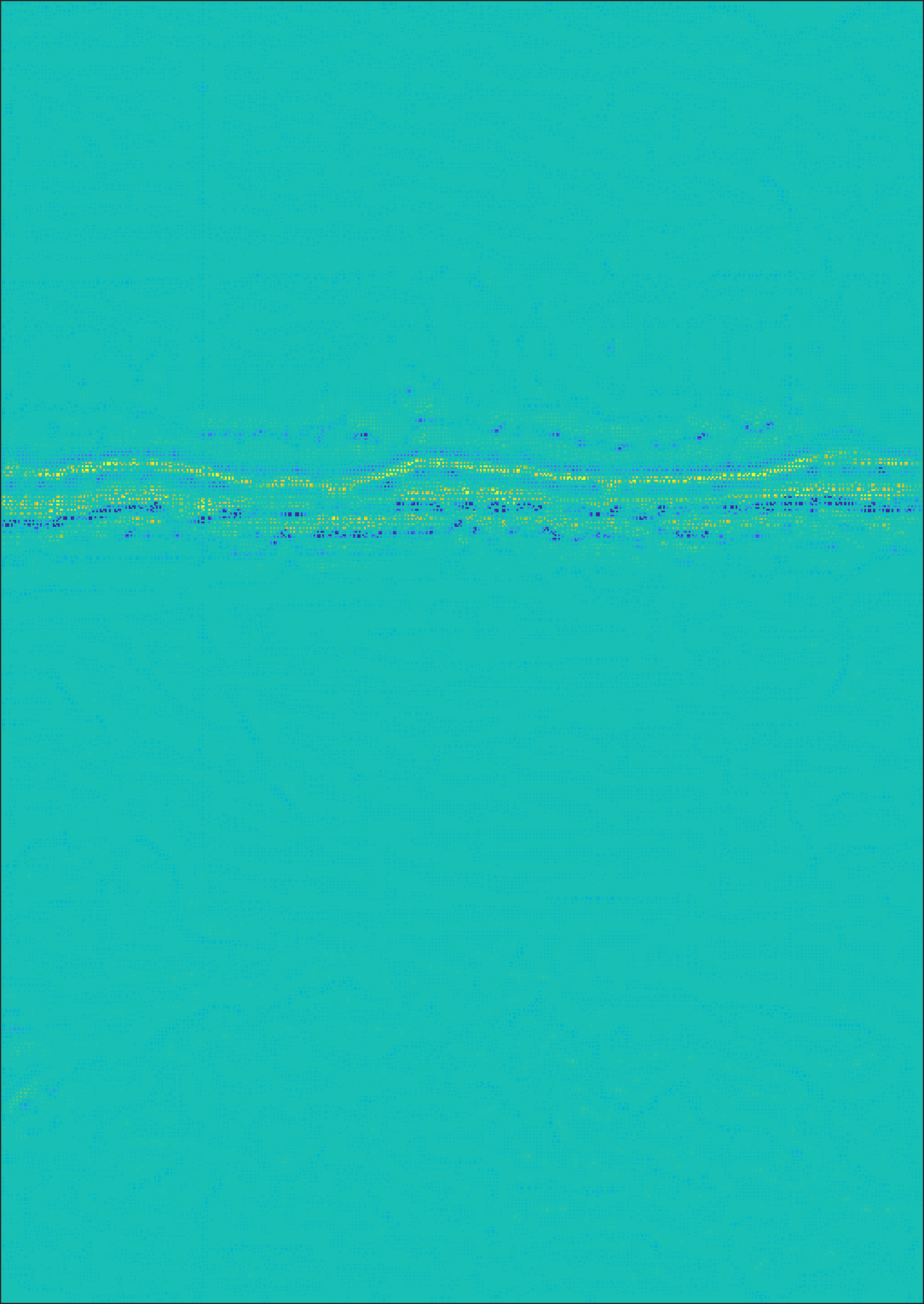}
   \caption{}
   \label{fig:8_1c}
\end{subfigure}
\begin{subfigure}[t]{0.18\textwidth}
   \includegraphics[width=1\linewidth]{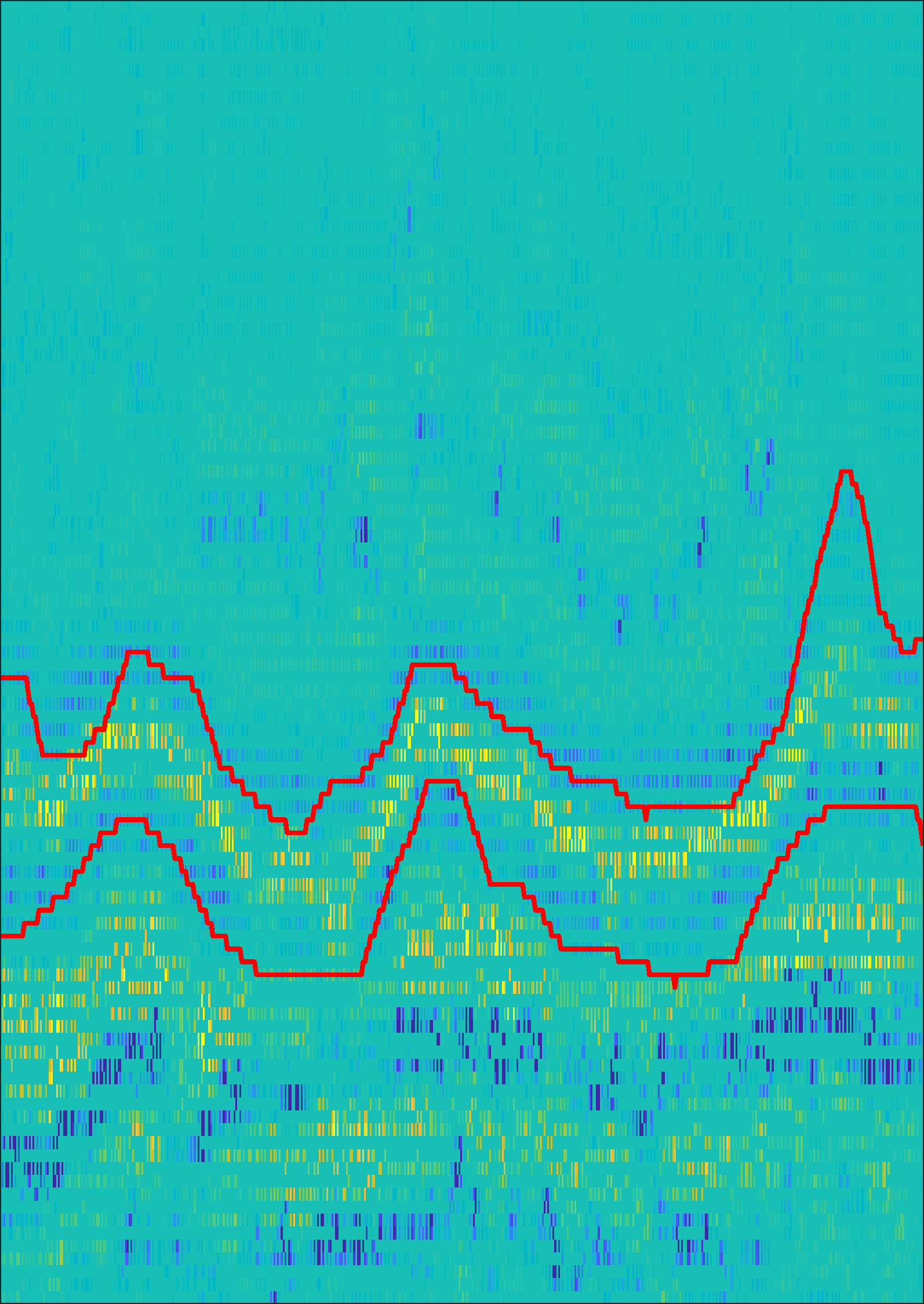}
   \caption{}
   \label{fig:8_1d}
\end{subfigure}
\\[\baselineskip]
\begin{subfigure}[t]{0.18\textwidth}
   \includegraphics[width=1\linewidth]{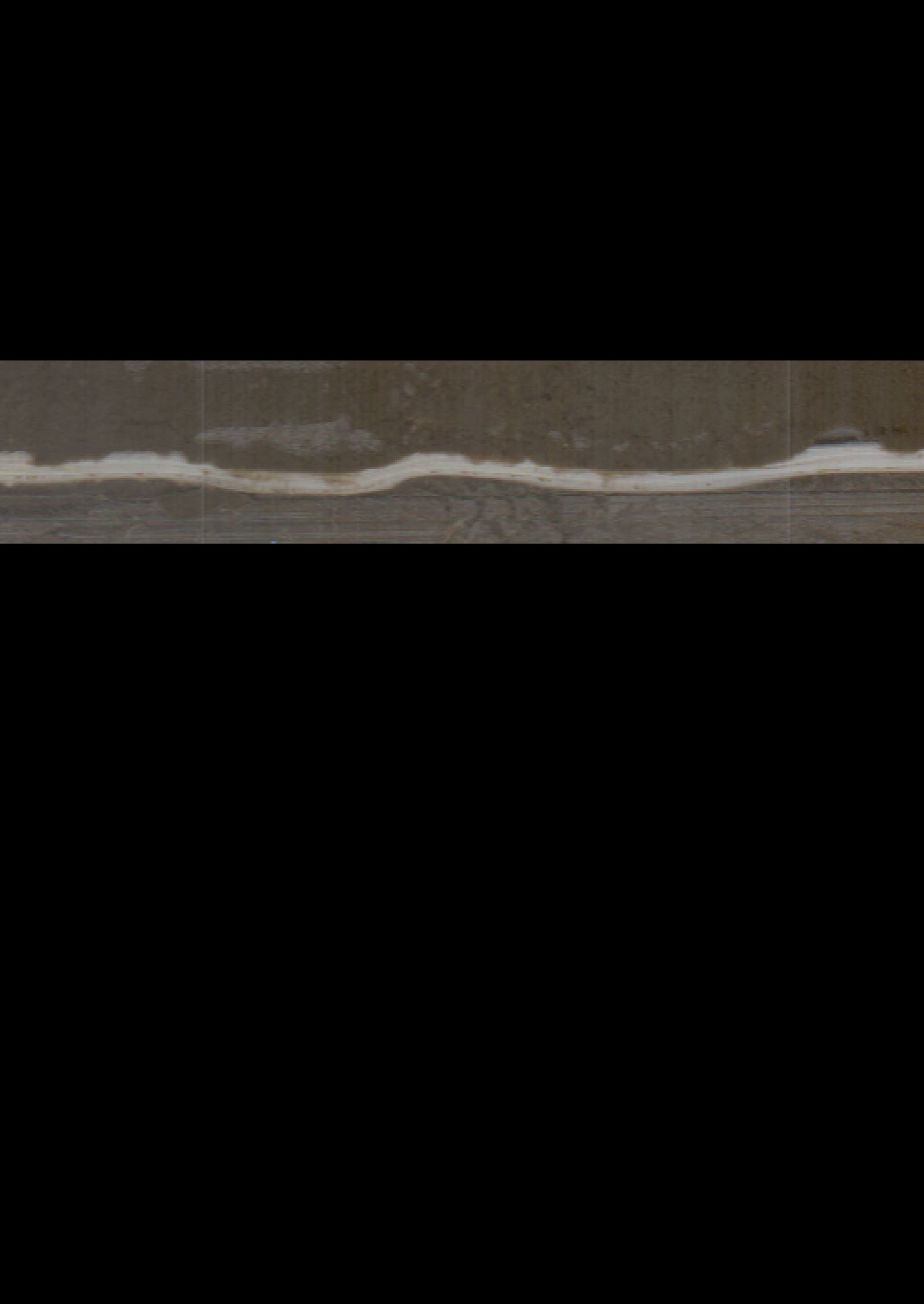}
   \caption{}
   \label{fig:8_1e}
\end{subfigure}
\begin{subfigure}[t]{0.18\textwidth}
   \includegraphics[width=1\linewidth]{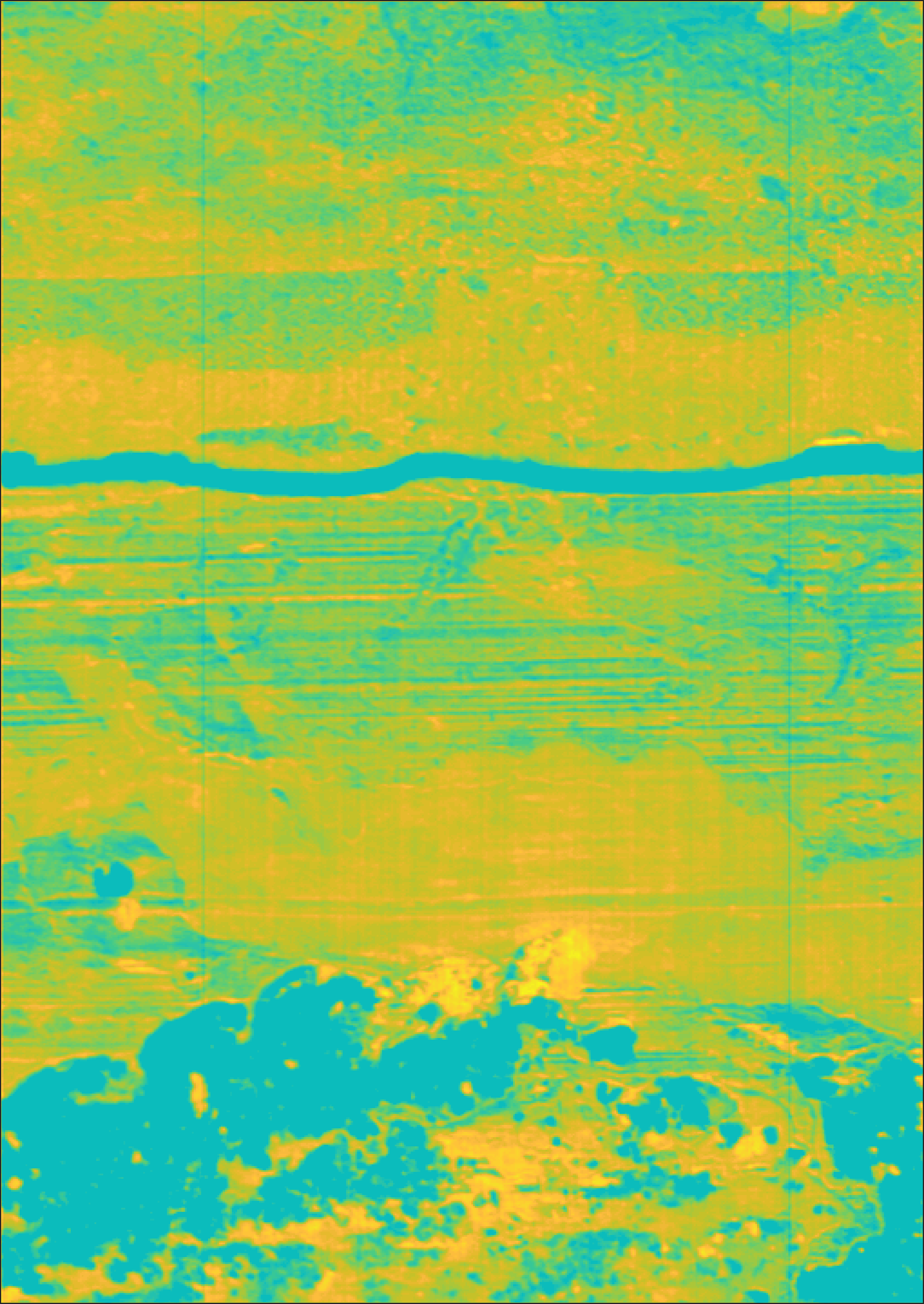}
  \caption{}
   \label{fig:8_1f}
\end{subfigure}
\begin{subfigure}[t]{0.18\textwidth}
   \includegraphics[width=1\linewidth]{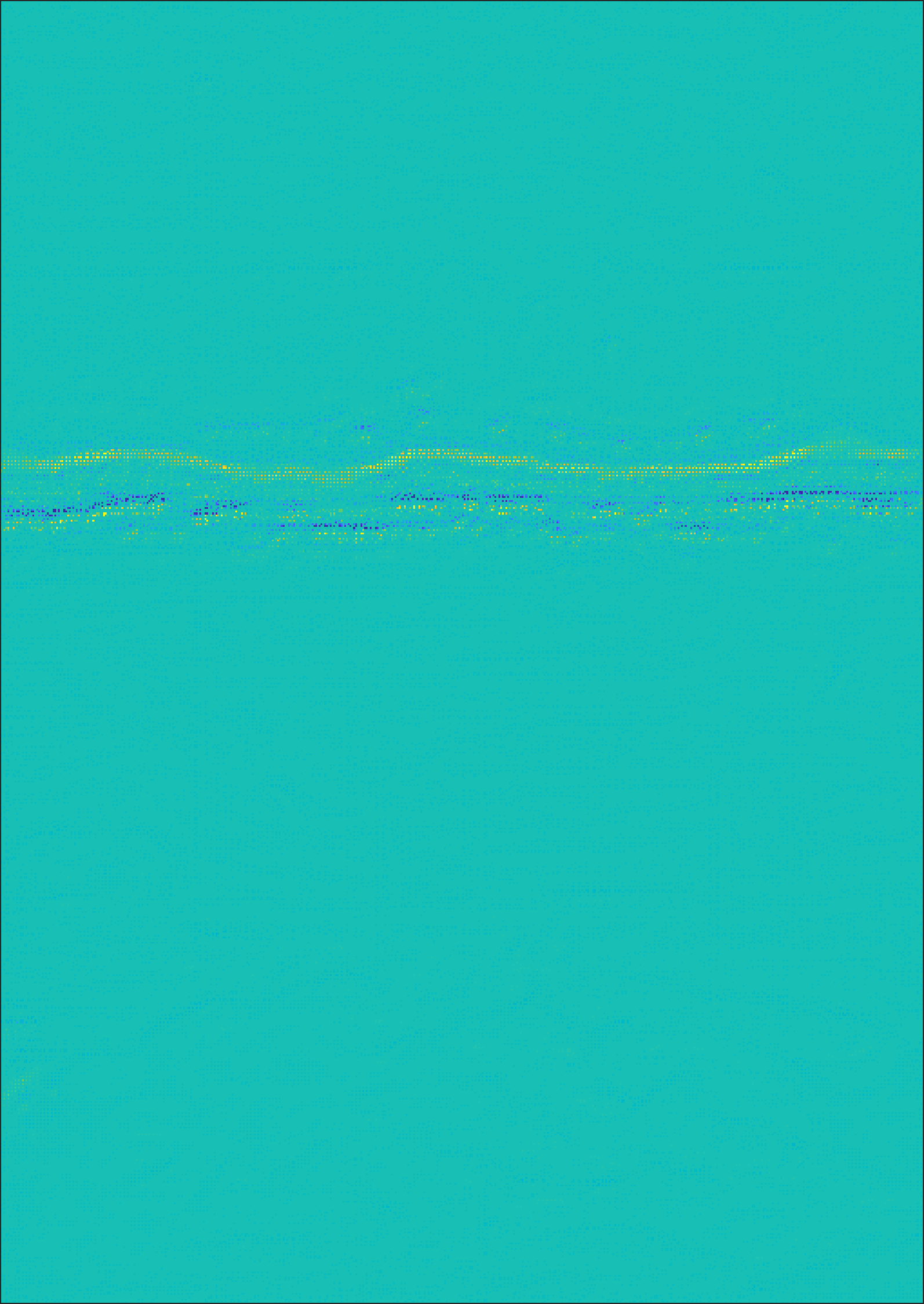}
   \caption{}
   \label{fig:8_1g}
\end{subfigure}
\begin{subfigure}[t]{0.18\textwidth}
   \includegraphics[width=1\linewidth]{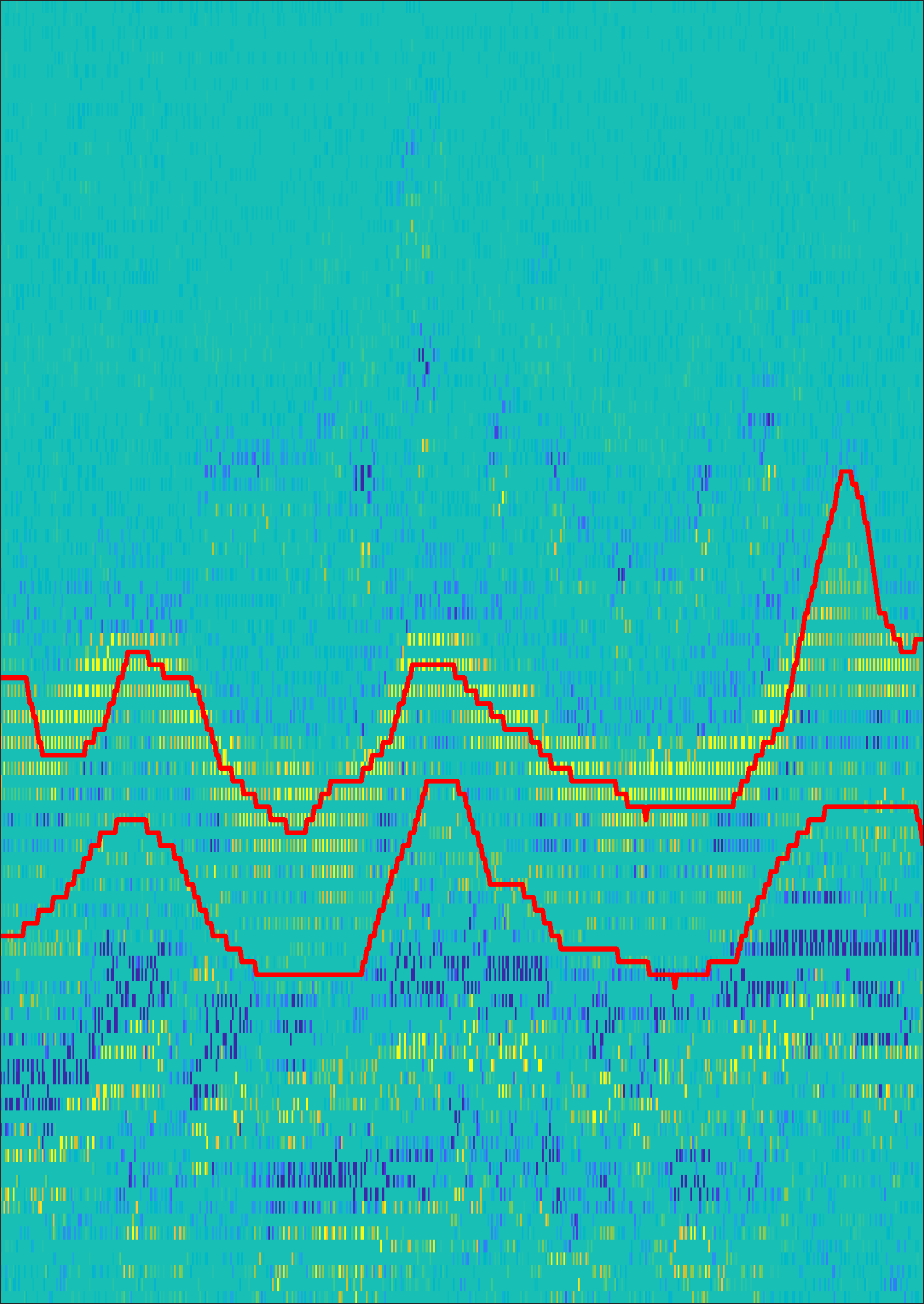}
  \caption{}
  \label{fig:8_1h}
\end{subfigure}
\begin{subfigure}[t]{0.058\textwidth}
   \includegraphics[width=1\linewidth]{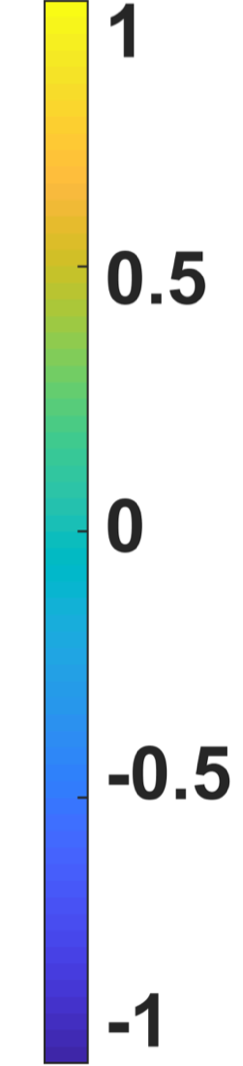}
  \caption{}
  \label{fig:8_1i}
\end{subfigure}

   \caption {Examples of the gradients with respect to the image classification score of the target root class using various individual feature maps. (\subref{fig:8_1a}) Original MR image. (\subref{fig:8_1b}) Feature map of Fig.\ref{fig:8_1a} which highlights the root area. (\subref{fig:8_1c}) Gradients of feature map in Fig.\ref{fig:8_1b} with respect to the root class score. (\subref{fig:8_1d}) Gradients in Fig.\ref{fig:8_1c} around the masked root regions shown in Fig.\ref{fig:8_1e}. The red lines indicate the boundary of root object. (\subref{fig:8_1e}) Cropped area from full image as mask in Fig.\ref{fig:8_1d} and Fig.\ref{fig:8_1h}. (\subref{fig:8_1f}) Feature map of Fig.\ref{fig:8_1a} which highlights soil. (\subref{fig:8_1g}) Gradients of feature map in Fig.\ref{fig:8_1g}. (\subref{fig:8_1h}) Gradients in Fig.\ref{fig:8_1g} around root object in Fig.\ref{fig:8_1e}. The red lines indicate the boundary of the root object. Fig.\ref{fig:8_1d} and Fig.\ref{fig:8_1h} are rescaled to match the size of other subfigure. (\subref{fig:8_1i}) Colorbar used in for images in Fig. 2.}
\label{fig:8_1}
\end{center}
\end{figure}
\subsection{Attention Maps for Semantic Segmentation}
CAM \cite{zhou2016learning} is one of the earliest methods showing that attention maps can localize the discriminative image components for classification.  CAM uses a network structure consisting of a block of fully convolutional layers followed by a global average pooling layer and a single fully connected layer. The block of fully convolutional layers extract image features. These extracted features are combined linearly using the weights of the final fully connected layer to define the CAM. However, the CAM from this approach often has a low resolution, making it challenging to infer pixel-level labels precisely for semantic segmentation. Grad-CAM \cite{selvaraju2017grad} is an extension of CAM which can estimate higher resolution attention maps by using features from any convolutional layer in the network. Specifically, Grad-CAM estimates the weights used for combining features as the average of gradients of the image classification score with respect to each value in the corresponding feature maps. Following the introduction of Grad-CAM, many methods were proposed to attempt to improve the quality of attention maps generated. Grad-CAM++ \cite{chattopadhay2018grad} takes a weighted average of only the positive gradients of the image classification score with respect to each feature map. SMOOTHGRAD \cite{smilkov2017smoothgrad} averages over several Grad-CAM estimated attention maps with added zero-mean Gaussian noise with the aim of reducing sensitivity to feature map noise. Smooth Grad-CAM++ \cite{omeiza2019smooth} mimics SMOOTHGRAD but applies the approach to Grad-CAM++ estimated attention maps.  Score-CAM \cite{wang2019score} attempts to improve Grad-CAM by weighting feature maps based on a metric which measures the increase of confidence for a class associated with the inclusion of each feature map. In application, all of these methods have been found to be either imprecise or sensitive to imbalanced data sets. Specifically in our application, soil pixels having complex gradients (i.e., both positive and negative gradients) which has a huge impact on the weights.   Consider the example shown in Fig.\ref{fig:8_1}.  Gradients across the feature maps have differing signs as shown in Fig.\ref{fig:8_1d} and Fig.\ref{fig:8_1h} and, thus, when averaged over the map may cancel each other out. Given this, the standard Grad-CAM approach is ineffective since the average of the gradients over the feature map is used to compute the attention map.

\subsection{Weakly Supervised Learning}
Weakly supervised and multiple instance learning (MIL) algorithms for image segmentation do not require precise pixel-level labels. Under MIL, a set of samples (e.g., an image) is labeled as either ``positive'' or ``negative.''  Positively labeled images are assumed to have at least one pixel corresponding to the target class (i.e., in our case, roots).  The number of target pixels in positively labeled images are unknown.  Negatively labeled images are composed of only non-target class (i.e., soil) pixels.  Often, MIL approaches iteratively estimate the likelihood each pixel is a target and, using these values, update classifier parameters (and, then, subsequently update likelihood values again) \cite{oquab2015object,durand2017wildcat}. Similarly, the pixels with the lowest target likelihood in each positively labeled image is also commonly assumed to be from the non-target class \cite{durand2017wildcat,durand2016weldon}. In contrast to methods that select likely target and non-target pixels, some methods have been proposed which consider all pixels in an image as equally contributing to the image-level label \cite{zhou2016learning}. The Log-Sum-Exp (LSE) algorithm uses a hyper-parameter which trades off between selecting a single pixel as the target representative and considering all pixels in an image as target with equal contribution \cite{pinheiro2015image}. Global weighted rank pooling (GWRP) is another way to generalize number of pixels identified as targets \cite{kolesnikov2016seed}. In all of these approaches, it is difficult to select a fixed number of pixels to identify as targets representatives.  

One reason that identifying the number of target pixels is challenging is that the size of the target class objects vary across images (e.g., some images contain only very few thin, fine roots whereas others are filled with roots of varying diameter). To alleviate this challenge, some approaches identify target pixels by adapting a threshold \cite{wei2018revisiting,huang2018weakly,ahn2019weakly,lee2019ficklenet}. In \cite{wei2018revisiting}, pixels with target class scores larger than a predefined threshold are labeled as targets and pixels with low saliency values are considered background. However, in this approach pixels are often unassigned to either target or background classes, pixels may be assigned to multiple target labels, or pixels may be assigned to background despite being surrounded in their neighborhood by target pixels.  In \cite{wei2017object}, pixels in all of these cases are ignored. The approach outlined in \cite{huang2018weakly} attempts to deal with these ignored pixels using deep seeded region growing (DSRG).  DSRG proposes to propagate labels from labeled pixels to unlabeled pixels. The method presented in \cite{lee2019ficklenet} extends the DSRG approach \cite{huang2018weakly} by thresholding aggregated localization maps to improve delineation of target regions and adapts their algorithm to accommodate semi-supervised segmentation. Following an initial segmentation, some approaches apply post-processing steps to smooth and improve segmentation labels.  These include conditional random fields (CRF) and the GWRP approach \cite{krahenbuhl2011efficient,ahn2019weakly,kolesnikov2016seed}. 

\subsection{MR Image Segmentation}
Several methods have been developed for automated minirhizotron image segmentation \cite{zeng2006detecting,heidari2014new,zeng2010rapid,rahmanzadeh2016novel}. Currently, supervised deep learning approaches are the methods that are achieving the state-of-art results in MR image segmentation \cite{xu2019overcoming,wang2019segroot,yasrab2019rootnav,smith2020segmentation}. Yet, deep learning methods require a large collection of precisely traced root images for training the networks.  A small number of approaches have been investigated for weakly supervised MR image segmentation \cite{yu2019root}. Yu, et al. \cite{yu2019root} studied the application of three MIL algorithms: multiple instance adaptive cosine coherence estimator (MI-ACE) \cite{zare2017discriminative}, multiple instance support vector machine (miSVM) \cite{andrews2003support}, and multiple instance learning with randomized trees (MIForests) \cite{leistner2010miforests} for application to MR imagery. These methods, however, did not do feature learning and, so, the authors manually identified color features to be used during segmentation.

\section{MIL-CAM Methodology} \label{sec3}
Semantic segmentation from weak labels using MIL-CAM is achieved in two training stages.  The first stage, outlined in Alg. \ref{alg:MILCAM}, estimates the set of parameters needed to compute an attention map $\textbf{S}^c \in \mathbb{R}^{M \times N} $ for a class $c$ where $M$ and $N$ are the numbers of rows and columns of the input image, respectively. The attention map is estimated using the softmax output of a weighted linear combination feature maps extracted from the various layers of a trained CNN as described in Eq. \ref{eq:1},

\begin{equation}
\mathbf{S}^c =  \displaystyle \frac{exp(\sum_{j} w^c_j y(\mathbf{F}_j) + b^c)}{\sum_{q}exp(\sum_{j} w^q_j y(\mathbf{F}_j) + b^q)}.
\label{eq:1}
\end{equation}
where $q$ is an index over all output classes,  $w^q_j \in \mathbb{R}$ is the weight estimated for class $q$ and feature map $\mathbf{F}_j \in \mathbb{R}^{A_j \times B_j}$, $A_j$ and $B_j$ are the number of rows and columns in the $j^{th}$ feature map, $b^q \in \mathbb{R}$ is an estimated bias term, and $y(\cdot)$ is an interpolation function to scale an input to the size of $M \times N$. 

Once attention maps are obtained using Alg. \ref{alg:MILCAM}, a segmentation network is then trained as outlined in Alg. \ref{alg:MILSeg}.  After training, the segmentation network maps input test imagery to get pixel level segmentation outputs. 

\subsection{Attention Map Estimation}

\begin{figure}[h] 
\begin{center}
   \includegraphics[width=1\linewidth]{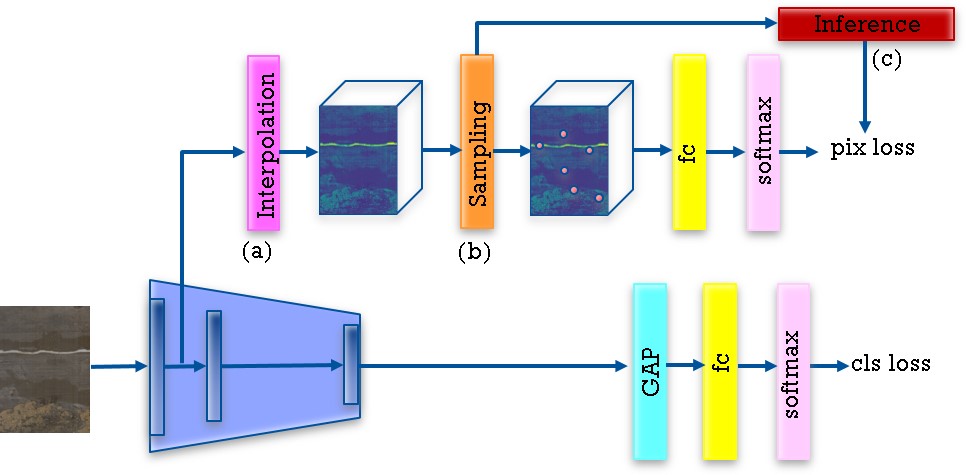}
   \caption{Architecture of MIL-CAM. GAP represents a global average pooling layer and fc represents a fully connected layer. cls loss represents the loss for image classification into positive (i.e., containing roots) or negative (i.e., does not contain roots). pix loss represents the loss for pixel level classification into root vs. soil.}
\label{fig:1_6}
\end{center}
\end{figure}

 MIL-CAM estimates the set of parameters needed to obtain attention maps and compute Eq. \ref{eq:1} using the combination of three key components: (a) a pixel-level feature extraction component; (b) a pixel sampling component used to form a bag for each image for MIL analysis; and (c) a linear model that performs the MIL-based segmentation. The sampled pixels with features extracted from the image classification network are used to train the linear model. The approach is illustrated in Fig. \ref{fig:1_6} and outlined in the following sub-sections. 

\subsubsection{Feature Extraction and Interpolation}
An image-level CNN classification network is first trained to extract coarse feature maps for each image. The training data set, $\{(\mathbf{I}_1, l_1),...(\mathbf{I}_k, l_k),...(\mathbf{I}_K, l_K) \}$, consists of $K$ images where each image $\mathbf{I}_k \in \mathbb{R}^{3 \times M \times N}$ is paired with image label $l_k \in \left\{0,1\right\}$ where 0 represents a negative image (i.e., does not contain roots) and 1 represents a positive image (i.e., contains roots). Using this training data an image-level classification network is trained by optimizing the cross-entropy loss as shown in Eq. 2,
\begin{equation}
\displaystyle \min_{\boldsymbol{\theta}_0} \displaystyle\sum_{k=1}^K L_{cls-loss}(\mathbf{I}_{k}; \boldsymbol{\theta}_0, \mathbf{l}_{k}) = \frac{-1}{K} \sum_{k=1}^K\sum_{q=1}^2l_{kq}\log {f_{q}(\mathbf{I}_{k};\boldsymbol{\theta}_0)} 
\label{eq:2}
\end{equation}
where $\mathbf{l}_{k} = [l_{k1}, l_{k2}]$ is the one-hot encoded label of the image $\mathbf{I}_{k}$ and  $f_{q}(\mathbf{I}_{k};\boldsymbol{\theta}_0)$ is the $q^{th}$ element of the softmax output layer of the the image classification network defined by parameters $\boldsymbol{\theta}_0$. The assumption is, provided an effective image-level classification network can be trained, that the network is extracting features that are useful for the semantic segmentation problem and these useful features are encoded in the CNN feature maps.  Once the classification network is trained, then the coarse CNN feature maps are upsampled using bilinear interpolation to match the size of the input image. Each pixel is represented by the corresponding feature vector obtained from the collection of upsampled feature maps.

\begin{algorithm2e}
\SetAlgoLined 
\KwData{$(\mathbf{I}, \mathbf{l}) = \{(\mathbf{I}_1, l_1),...(\mathbf{I}_k, l_k),...(\mathbf{I}_K, l_K) \}$}
Train the image classification network with $(\mathbf{I}, \mathbf{l})$ \;
Extract feature maps $\mathbf{F}$ from image classification network for $(\mathbf{I}, \mathbf{l})$  \;
Interpolate the CNN feature maps $y(\mathbf{F})$ for $(\mathbf{I}, \mathbf{l})$\;
Sample instances and construct bags, $\{(\mathbf{B}_1, l_1),...(\mathbf{B}_k, l_k),...(\mathbf{B}_K, l_K)\}$\;

Initialize each instance label with the label of its corresponding bag\;
\Repeat{Fixed number of epochs completed}{
 Update $\mathbf{w},\mathbf{b}$ by optimizing Eq. \ref{eq:4} with stochastic gradient descent for one epoch using the instances and updated labels ${(\mathbf{x}_{k}^{n}, \mathbf{l}_{k}^{n})}$\;
 Compute $p_{k}^{n} = g(\mathbf{x}_{k}^n;\mathbf{w},\mathbf{b})$ for each instance\;
 \For{Every positive bag $(\mathbf{B}_k, l_k=1)$}{
    $p_{t}$ = Otsu's($\{p_{k}^{1},...p_{k}^{N_{k}}\}$)\;
    If $p_{k}^{n} \geq p_{t} $, then set $\mathbf{l}_{k}^{n}$ as target, else set $\mathbf{l}_{k}^{n}$ as non-target\;}
 \For{Every negative bag $(\mathbf{B}_k, l_k=0)$}{
set $\mathbf{l}_{k}^{n}$ as non-target\;
}
}
\Return{$\boldsymbol{\theta}_0$,$\mathbf{w}, \mathbf{b}$ from epoch with smallest loss}
\caption{Estimating Weights and Biases for Attention Maps}
\label{alg:MILCAM}
\end{algorithm2e}

\subsubsection{Instance Sampling}
In order to address some of the imbalance in the data set (i.e., there are many more soil pixels than root pixels), a sampling approach is used to identify representative pixels from each image. The green band of the RGB minirhizotron image is used for instance sampling. The approach draws a single pixel to represent the set of pixels from each possible 8-bit value from the green band in the image. In other words, a 256 bin histogram is built using the values of the green band of the MR imagery. For each non-empty bin, a uniform random draw is used to identify a representative pixel for that green-level. In our application, we found this to be an effective approach to re-balance root-vs-nonroot pixels in positively labeled imagery (given that pixel level labels are unavailable).  The sampled pixels are organized into a set of bags, $\{(\mathbf{B}_1, l_1),...(\mathbf{B}_k, l_k),...(\mathbf{B}_K, l_K) \}$. Each bag, $\mathbf{B}_{k} = \left\{\mathbf{x}_{k}^{1},\mathbf{x}_{k}^{2},\dotsc,\mathbf{x}_{k}^{N_{k}}\right\}$, corresponds to one image $\mathbf{I}_k$ with image label $l_K$ and is composed of $N_k$ instances.  The instance $\mathbf{x}_{k}^{n} \in \mathbb{R}^{J}$ is the feature vector for the $n^{th}$ instance in the $k^{th}$ bag where $J$ is the number of feature maps used to construct the feature vectors. 

\begin{algorithm2e}
\SetAlgoLined 
\KwData{$ \{(\mathbf{I}_1, l_1),...(\mathbf{I}_k, l_k),...(\mathbf{I}_K, l_K) \}$}
\KwPara{$s_{t}$}

\For{Every positive image}{
Compute the score-map $\mathbf{S}_k^c$ from MIL-CAM using $(\boldsymbol{\theta}_0,\mathbf{w},\mathbf{b})$ in Eq.\ref{eq:1} \;
Estimate a threshold, $o_{t}$ = Otsu's $(\mathbf{S}_k^c)$\;
If  $\mathbf{S}_k^c(m,n) \geq o_{t} $, then set $\mathbf{l}_{k}^0(m,n)$ as target, else set $\mathbf{l}_{k}^0(m,n)$ as non-target\;}
\For{Every negative image $(\mathbf{I}_k, l_k=0)$}{  
set $\mathbf{l}_{k}^0(m,n)$ as non-target\;
}
Update parameters $\boldsymbol{\theta}_1$ for data set with pixel labels $\mathbf{l}_{k}^0(m,n)$ for a fixed number of epochs\;
\Repeat{Fixed number of epochs completed}{
   Compute score-map for each image using the U-Net with updated parameters\;
   \uIf{A positive image $(\mathbf{I}_k, l_k=1)$}{
   If $\mathbf{P}_{k}(m,n) \geq s_{t} $, then set $\mathbf{l}_{k}(m,n) $ as target, else $\mathbf{l}_{k}(m,n)$ as non-target\; 
   If every pixel $\mathbf{P}_{k}(m,n) < s_{t}$ , then $\mathbf{l}_{k}(m,n) = \mathbf{l}_{k}^0(m,n)$\;
   }
   \uElseIf{A negative image $(\mathbf{I}_k, l_k=0)$}{
  set $\mathbf{l}_{k}(m,n) $ as non-target\;
}
  Update parameters $\boldsymbol{\theta}_1$ for dataset with pixel labels $\mathbf{l}_{k}(m,n)$ \;
}
\Return{Segmentation network parameters $\boldsymbol{\theta}_1$} 
\caption{Weakly Supervised Image Segmentation}
\label{alg:MILSeg}
\end{algorithm2e}

\subsubsection{Estimated Weights and Biases}
After instance sampling, the weights and biases used to compute the attention maps as defined in Eq. \ref{eq:1} are estimated by optimizing the cross-entropy loss shown in Eq. \ref{eq:4} given the MIL constraints that for each positive bag, at least one instance must be labeled as root and all instances in every negative bag are labeled as non-root,
\begin{equation}
\displaystyle \min_{\mathbf{l}_{k}^{n}} \min_{\mathbf{w},\mathbf{b}} 
\displaystyle\sum_{\mathbf{x}_{k}^{n}} L_{pix-loss}( \mathbf{x}_{k}^{n};\mathbf{w},\mathbf{b}, \mathbf{l}_{k}^{n}) = \frac{-1}{\sum_{k}{N_k}}\sum_{\mathbf{x}_{k}^{n}}\sum_{q}l_{kq}^n\log {g_{q}(\mathbf{x}_{k}^n;\mathbf{w},\mathbf{b})} 
\label{eq:4}
\end{equation}
where $\mathbf{l}_{k}^n$ is the one-hot encoded label of the instance $\mathbf{x}_{k}^{n}$ and $g_{q}(\mathbf{x}_{k}^n;\mathbf{w},\mathbf{b})$ is the $q^{th}$ element of the softmax output of the MIL-CAM with parameters $(\mathbf{w},\mathbf{b})$. The loss is updated iteratively as outlined in Alg. \ref{alg:MILCAM}. During the initial epoch, each instance is labeled the same label as its bag. In all subsequent epochs, the probability that an instance belongs to the target class, $p_{k}^{n} = g(\mathbf{x}_{k}^n;\mathbf{w},\mathbf{b})$, is predicted by the linear model trained from the previous epoch. Then for each positive bag, a threshold $p_{t}$ is computed using Otsu's threshold \cite{otsu1979threshold} and all instances greater than the threshold are labeled as target whereas all others are labeled as non-target. For negative bags, all instances are labeled as non-target.

\subsection{Training the Image Segmentation Network}
\begin{figure}[h] 
\begin{center}
   \includegraphics[width=1\linewidth]{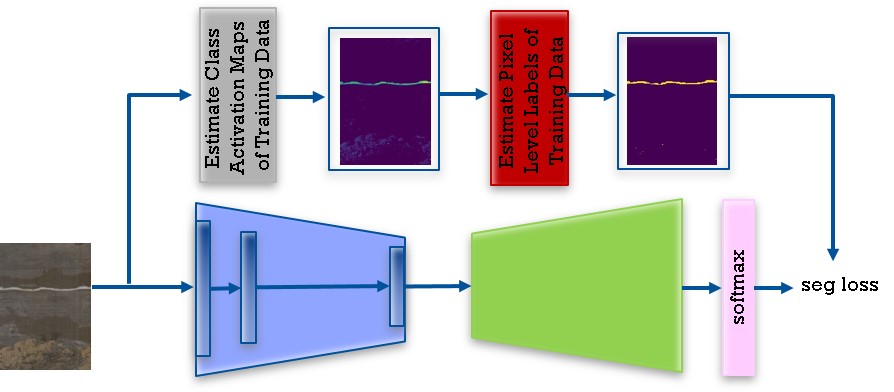}
   \caption{Architecture of segmentation U-Net with MIL training branch. The bottom branch is the U-Net. The top branch is used to infer label of training data.}
\label{fig:1_5}
\end{center}
\end{figure}

Once MIL-CAM attention maps can be estimated, an image segmentation network is trained as outlined in Alg.\ref{alg:MILSeg}. First,  target class attention maps for positively labeled images are estimated and thresholded using Otsu's threshold to obtain a label for each pixel.  All pixels in negatively labeled images are given a non-target label.  These labels are used to estimate the parameters for the U-Net \cite{ronneberger2015u} architecture illustrated in lower branch of Fig. \ref{fig:1_5}.  After initially training the U-Net with labels obtained from the attetnion maps, the U-Net is iteratively fine-tuned.  A score-map, $\mathbf{P}_k \in \mathbb{R}^{M \times N}$, is computed using the soft-max output of the U-Net.  The score-map of positively-labeled is thresholded using a fixed (large) threshold parameter, $s_t$, to obtain updated pixel level labels which highlight more likely positive samples. The updated labels are iteratively used to fine-tune the parameters of the U-Net. 

\section{Experiments} \label{sec4}
\subsection{Data Description}
For our experiments, we used a switchgrass (\textit{Panicum virgatum} L.) MR imagery dataset consisting of 561 training images with image-level labels and 30 test and validation images with pixel-level labels. Each image was $2160 \times 2550$ in size and was divided into sub-images of size $720 \times 510$. 500 sub-images containing roots and 500 sub-images containing only soil were randomly selected as training data for estimating attention map parameters. 1500 root sub-images and 1500 soil sub-images were randomly selected as training data for the U-Net segmentation network. The 30 images with pixel-level labels were randomly divided into 10 validation images and 20 test images.   
 
\subsection{Architecture}
Our experiments use U-Net \cite{ronneberger2015u} with layer depth of 5 as backbone for MR image segmentation. The feature extraction network used to estimate attention map parameters was a 2-class convolutional neural network with the encoder of the U-Net, followed by a global average pooling layer and a fully connected layer. We extract $1024\times 46 \times 33$ feature maps and vectorize the feature maps to classify each image into 2 classes with a fully connected layer.  The feature extraction net is trained using SGD at a learning rate of 0.0001 and momentum of 0.8 in the online mode to minimize the cross entropy loss. The MIL-CAM attention map module extracts a 64-dimensional feature for each sampled instance from the fourth layer of the encoder of the feature extraction network. Then, classifies each sampled instance into one of two classes using a fully connected layer. The MIL-CAM attention map module is trained using SGD at a learning rate of 0.001 and momentum of 0.5 in the online mode to minimize the cross entropy loss. 

The image segmentation network was a U-Net of depth 5 and a MIL training branch. The MIL training branch extracts $64 \times 720 \times510$ features from the first layer of the encoder of the feature extraction network and compute a $720 \times 510$ score-map of target class for each training image.  The threshold parameter $s_t$ was set to 0.9 to estimate pixel label from the score-map.  The U-Net was first initialized for 10 epochs using Adam at learning rate of 0.0001 in the online mode to minimize the cross entropy loss where the root class was weighted by 50 using the labels produced by the attention maps. Then, during iterative fine-tuning, the network parameters were also updated using Adam with learning rate of 0.0001 in the online mode to minimize the cross entropy loss with the root class having an additional weight of 50. The weight on root class addressed the imbalance issue between root class and soil class.

\subsection{Experiments: MIL-CAM Attention Maps}

\begin{figure}[h] 
\begin{center}
\begin{subfigure}[t]{0.15\textwidth}
   \includegraphics[width=1\linewidth]{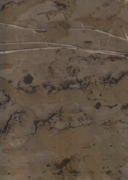}
\end{subfigure}
\begin{subfigure}[t]{0.15\textwidth}
   \includegraphics[width=1\linewidth]{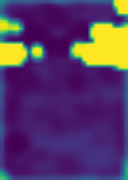}
\end{subfigure}
\begin{subfigure}[t]{0.15\textwidth}
   \includegraphics[width=1\linewidth]{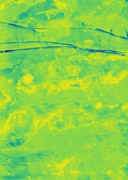}
\end{subfigure}
\begin{subfigure}[t]{0.15\textwidth}
   \includegraphics[width=1\linewidth]{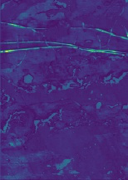}
\end{subfigure}
\begin{subfigure}[t]{0.15\textwidth}
   \includegraphics[width=1\linewidth]{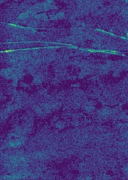}
\end{subfigure}
\begin{subfigure}[t]{0.15\textwidth}
   \includegraphics[width=1\linewidth]{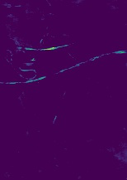}
\end{subfigure}
\\[\baselineskip]
\begin{subfigure}[t]{0.15\textwidth}
   \includegraphics[width=1\linewidth]{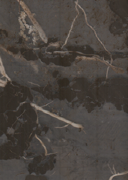}
\end{subfigure}
\begin{subfigure}[t]{0.15\textwidth}
   \includegraphics[width=1\linewidth]{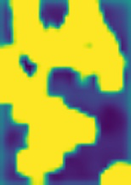}
\end{subfigure}
\begin{subfigure}[t]{0.15\textwidth}
   \includegraphics[width=1\linewidth]{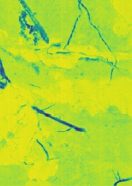}
\end{subfigure}
\begin{subfigure}[t]{0.15\textwidth}
   \includegraphics[width=1\linewidth]{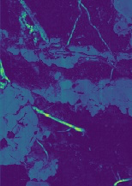}
\end{subfigure}
\begin{subfigure}[t]{0.15\textwidth}
   \includegraphics[width=1\linewidth]{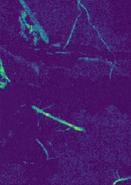}
\end{subfigure}
\begin{subfigure}[t]{0.15\textwidth}
   \includegraphics[width=1\linewidth]{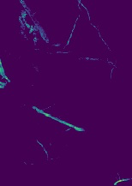}
\end{subfigure}
\\[\baselineskip]
\begin{subfigure}[t]{0.15\textwidth}
   \includegraphics[width=1\linewidth]{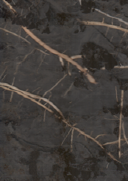}
   \caption{}
   \label{fig:3_1a}
\end{subfigure}
\begin{subfigure}[t]{0.15\textwidth}
   \includegraphics[width=1\linewidth]{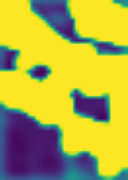}
   \caption{}
   \label{fig:3_2b}
\end{subfigure}
\begin{subfigure}[t]{0.15\textwidth}
   \includegraphics[width=1\linewidth]{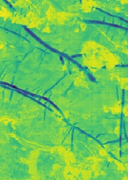}
   \caption{}
   \label{fig:3_3c}
\end{subfigure}
\begin{subfigure}[t]{0.15\textwidth}
   \includegraphics[width=1\linewidth]{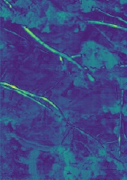}
   \caption{}
   \label{fig:3_4d}
\end{subfigure}
\begin{subfigure}[t]{0.15\textwidth}
   \includegraphics[width=1\linewidth]{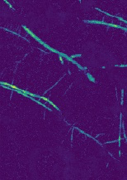}
   \caption{}
   \label{fig:3_5e}
\end{subfigure}
\begin{subfigure}[t]{0.15\textwidth}
   \includegraphics[width=1\linewidth]{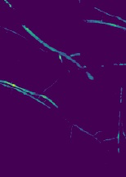}
   \caption{}
   \label{fig:3_6f}
\end{subfigure}

   \caption{Attention maps of different methods. (\subref{fig:3_1a}) Original Image. (\subref{fig:3_2b}) Result of CAM. (\subref{fig:3_3c}) Result of Grad-CAM. (\subref{fig:3_4d}) Result of Grad-CAM++. (\subref{fig:3_5e}) Result of SMOOTHGRAD. (\subref{fig:3_6f}) Result of MIL-CAM.}
\label{fig:3_1}
\end{center}
\end{figure}
The attention maps of MIL-CAM were first qualitatively compared with attention maps of other methods as shown in Fig.\ref{fig:3_1}. As can be seen, MIL-CAM results shown in Fig.\ref{fig:3_6f} more accurately indicate root locations as compared to the attention maps produced by CAM in Fig.\ref{fig:3_2b}. This difference in performance is largely due to the fact that CAM requires interpolating a low resolution attention map to the size of the input image resulting in blurred, oversized detection regions. Grad-CAM in Fig.\ref{fig:3_3c}. fails to correctly identify roots and, instead, highlights soil. Furthermore, MIL-CAM produced attention maps with higher contrast between root pixels and background than those Grad-CAM ++ in Fig.\ref{fig:3_4d} and SMOOTHGRAD in Fig.\ref{fig:3_5e}.

\begin{figure}[h] 
\begin{center}
\begin{subfigure}[t]{0.15\textwidth}
   \includegraphics[width=1\linewidth]{Fig3_1_6_org.png}
\end{subfigure}
\begin{subfigure}[t]{0.15\textwidth}
   \includegraphics[width=1\linewidth]{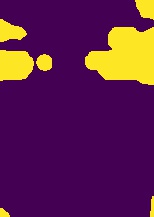}
\end{subfigure}
\begin{subfigure}[t]{0.15\textwidth}
   \includegraphics[width=1\linewidth]{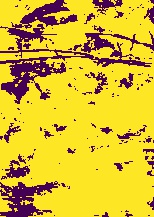}
\end{subfigure}
\begin{subfigure}[t]{0.15\textwidth}
   \includegraphics[width=1\linewidth]{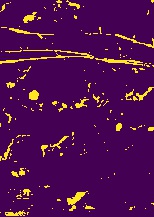}
\end{subfigure}
\begin{subfigure}[t]{0.15\textwidth}
   \includegraphics[width=1\linewidth]{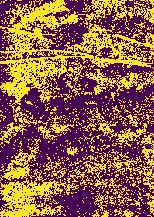}
\end{subfigure}
\begin{subfigure}[t]{0.15\textwidth}
   \includegraphics[width=1\linewidth]{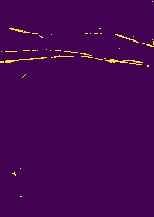}
\end{subfigure}
\\[\baselineskip]
\begin{subfigure}[t]{0.15\textwidth}
   \includegraphics[width=1\linewidth]{Fig3_2_6_org.png}
\end{subfigure}
\begin{subfigure}[t]{0.15\textwidth}
   \includegraphics[width=1\linewidth]{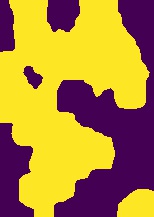}
\end{subfigure}
\begin{subfigure}[t]{0.15\textwidth}
   \includegraphics[width=1\linewidth]{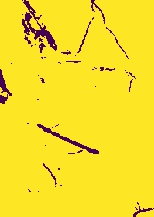}
\end{subfigure}
\begin{subfigure}[t]{0.15\textwidth}
   \includegraphics[width=1\linewidth]{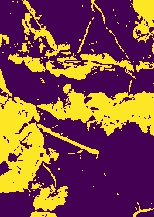}
\end{subfigure}
\begin{subfigure}[t]{0.15\textwidth}
   \includegraphics[width=1\linewidth]{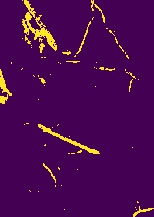}
\end{subfigure}
\begin{subfigure}[t]{0.15\textwidth}
   \includegraphics[width=1\linewidth]{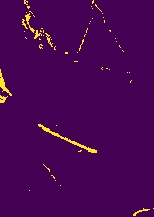}
\end{subfigure}
\\[\baselineskip]
\begin{subfigure}[t]{0.15\textwidth}
   \includegraphics[width=1\linewidth]{Fig3_3_6_org.png}
   \caption{}
   \label{fig:7_1a}
\end{subfigure}   
\begin{subfigure}[t]{0.15\textwidth}
   \includegraphics[width=1\linewidth]{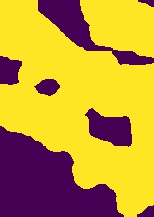}
   \caption{}
   \label{fig:7_2b}
\end{subfigure}
\begin{subfigure}[t]{0.15\textwidth}
   \includegraphics[width=1\linewidth]{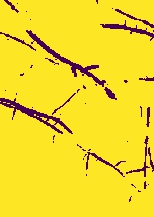}
   \caption{}
   \label{fig:7_3c}
\end{subfigure}
\begin{subfigure}[t]{0.15\textwidth}
   \includegraphics[width=1\linewidth]{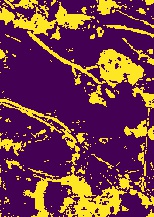}
   \caption{}
   \label{fig:7_4d}
\end{subfigure}
\begin{subfigure}[t]{0.15\textwidth}
   \includegraphics[width=1\linewidth]{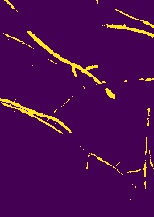}
   \caption{}
   \label{fig:7_5e}
\end{subfigure}
\begin{subfigure}[t]{0.15\textwidth}
   \includegraphics[width=1\linewidth]{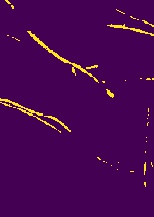}
   \caption{}
   \label{fig:7_6f}
\end{subfigure}

   \caption{Thresholded attention maps. (\subref{fig:7_1a}) Original Image. (\subref{fig:7_2b}) Result of CAM. (\subref{fig:7_3c}) Result of Grad-CAM. (\subref{fig:7_4d}) Result of Grad-CAM++. (\subref{fig:7_5e}) Result of SMOOTHGRAD. (\subref{fig:7_6f}) Result of MIL-CAM.}

\label{fig:7_1}
\end{center}
\end{figure}

\begin{table}[h]
\begin{center}
\caption{Compare results of thresholded attention maps}
\label{table:inference}
\begin{tabular}{|l|*{11}{c|}}
\hline
Method & Precision & Recall & F1 score & mIoU \\
\hline
CAM & $0.045 \pm0.0053$ & $ \textbf{0.931} \pm \textbf{0.0459}$ & $0.085 \pm 0.0098$ & $0.045 \pm0.0053$\\
\hline
Grad-CAM & $0.003\pm0.0012$ & $0.229\pm0.0939$ & $0.006\pm0.0024 $ & $0.003\pm0.0012 $\\
\hline
Grad-CAM++ & $0.015 \pm 0.0084$ & $0.550\pm0.1951$ & $0.030\pm0.0159$ & $0.015\pm0.0083$\\
\hline
SMOOTHGRAD&$0.033 \pm 0.0028$ & $0.782 \pm 0.0191$ & $0.064 \pm 0.0052$ & $0.033 \pm 0.0028$\\
\hline
MIL-CAM & $\textbf{0.248} \pm \textbf{0.1870}$ & $0.536\pm0.1450$ & $\textbf{0.289}\pm \textbf{0.1814} $ & $\textbf{0.177}\pm \textbf{0.1190}$\\
\hline
\end{tabular}
\end{center}
\end{table}

Fig.\ref{fig:7_1} compares attention maps from a selection of approaches after thresholding with Otsu's threshold. Table \ref{table:inference} lists the average and standard devation for precision, recall and F1 score of three training runs of the various approaches to compare the quality these thresholded results. The proposed MIL-CAM method has a significantly higher F1 score among all those compared. The precision of MIL-CAM is an order of magnitude better than the comparison methods without a significant loss in recall as compared with the gains of  precision. Although other methods except Grad-CAM have a better recall, the low precision scores of these methods indicate a large amount of background pixels are mislabeled as root pixels.  This can be visualized in Fig.\ref{fig:7_1}. 

\subsection{Experiments: Semantic Segmentation}

\begin{table*}[h]
\begin{center}
\caption{Comparison of image segmentation results. All comparison methods use weak image level labels except the U-Net approach from \cite{xu2019overcoming}. MIL-CAM Th is the result found after thresholding the U-Net softmax outputs corresponding to the target class at FPR = 0.03; argmax MIL-CAM is the result when taking the argmax of U-Net softmax outputs; and MIL-CAM + CRF method is the result when the argmax MIL-CAM result is postprocessed with a CRF.}
\label{table:post}
\begin{tabular}{|l|*{11}{c|}}
\hline
Method & Label & Precision & Recall & F1 score & mIoU   \\
\hline
U-Net \cite{xu2019overcoming} & pixel &$0.307$ &$\textbf{0.913}$&$0.459$ & $0.298$\\
\hline
\hline
MI-ACE\cite{yu2019root} & image & $0.130\pm 0.0010$ & $0.775\pm0.0067$ & $0.223\pm0.0017$ & $0.125\pm0.0011$  \\
\hline
miSVM\cite{yu2019root} & image & $0.134\pm0.0015$ & $0.798\pm0.0104$ & $0.229\pm0.0026$ & $0.129\pm0.0017$ \\
\hline
MIForests\cite{yu2019root} & image & $0.101\pm0.0104$ & $0.582\pm0.0664$ & $0.172\pm0.0180$ & $0.094\pm0.0108$   \\
\hline
MIL-CAM Th & image & $0.145\pm0.0050 $ & $ 0.878\pm0.0341$ & $0.249\pm0.0088$ & $0.142\pm0.0057$  \\
\hline
\hline
argmax MIL-CAM & image & $0.186\pm 0.0278$ &$0.859\pm0.0423$ &$0.304\pm0.0364$ & $0.180\pm0.0251$\\
\hline
MIL-CAM + CRF & image & $\textbf{0.667} \pm \textbf{0.0257}$ &$0.692 \pm 0.0267$&$\textbf{0.678}\pm \textbf{0.0058}$& $\textbf{0.513} \pm \textbf{0.0066}$ \\
\hline
\end{tabular}
\end{center}
\end{table*}

We also compared the performance of our final MIL segmentation network (i.e., MIL-CAM Th in the table) against other MIL methods (MI-ACE\cite{yu2019root}, miSVM\cite{yu2019root}, and MIForest\cite{yu2019root}). The average and standard deviation of three runs of the precision, recall, F1 score and mIoU were compared at false positive rate (FPR) is 0.03 in Table \ref{table:post}.  Our proposed approach outperformed all other MIL methods. The proposed MIL-CAM Th method (i.e., the thresholded MIL-CAM result) achieved recall$= 0.878$. The recall of MIL-CAM Th was 10\% better than miSVM which was the second best.  MIL-CAM Th also had the best precision of all MIL methods.

\begin{figure*}[h] 
\begin{center}
\begin{subfigure}{0.17\textwidth}
   \includegraphics[width=1\linewidth]{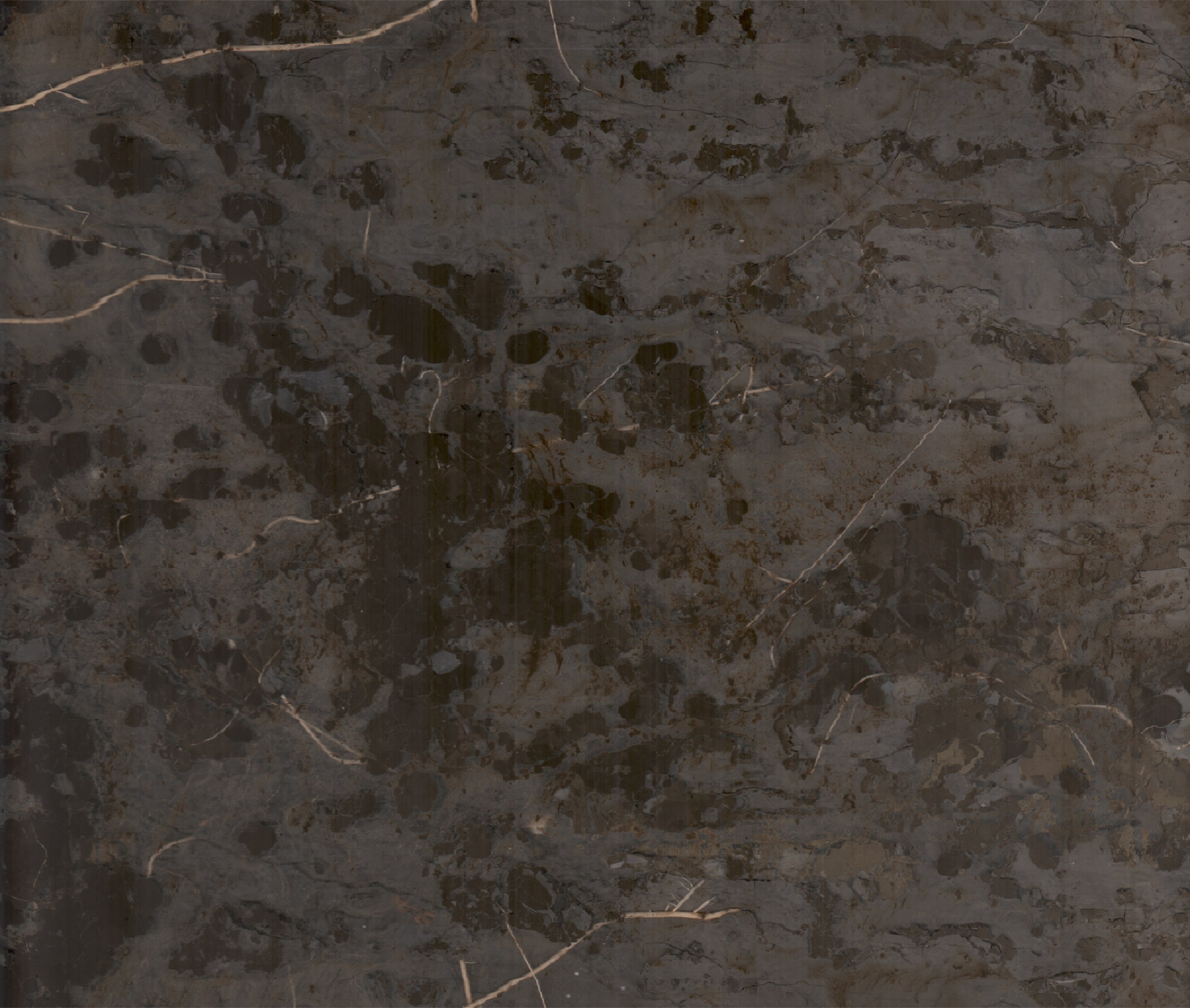}
\end{subfigure}
\begin{subfigure}{0.17\textwidth}
   \includegraphics[width=1\linewidth]{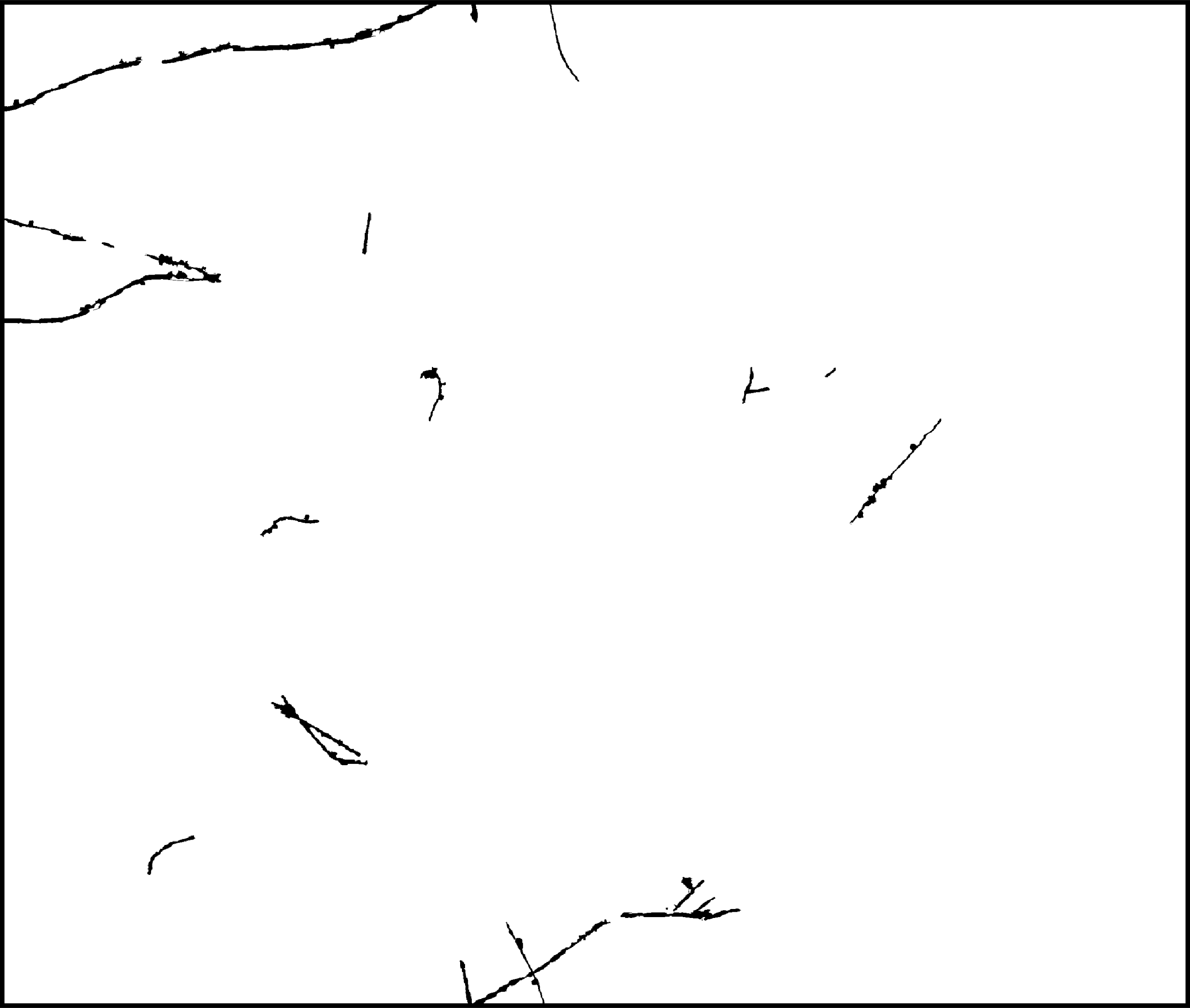}
   \end{subfigure}
\begin{subfigure}{0.17\textwidth}
   \includegraphics[width=1\linewidth]{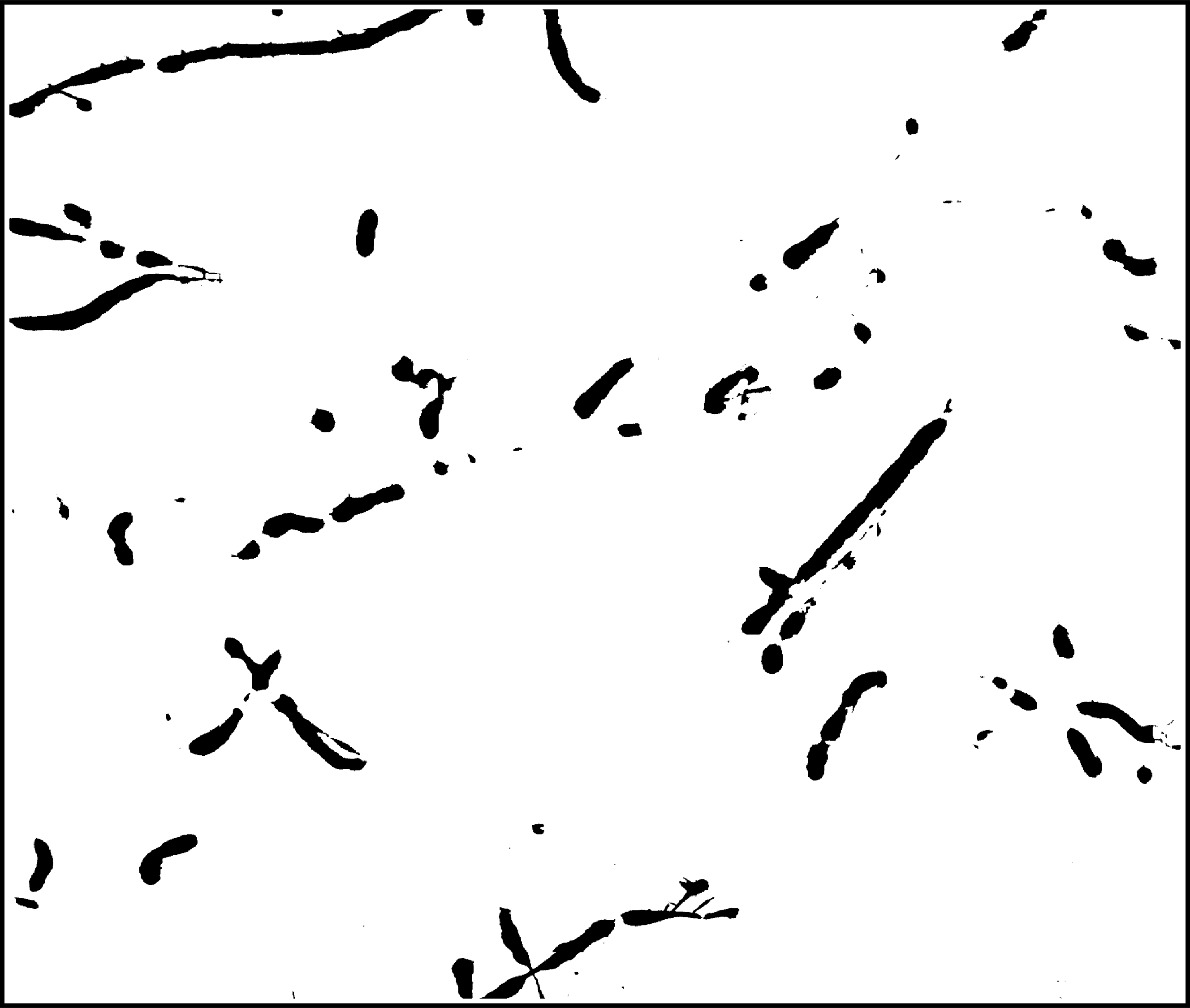}
   \end{subfigure}
\begin{subfigure}{0.17\textwidth}
   \includegraphics[width=1\linewidth]{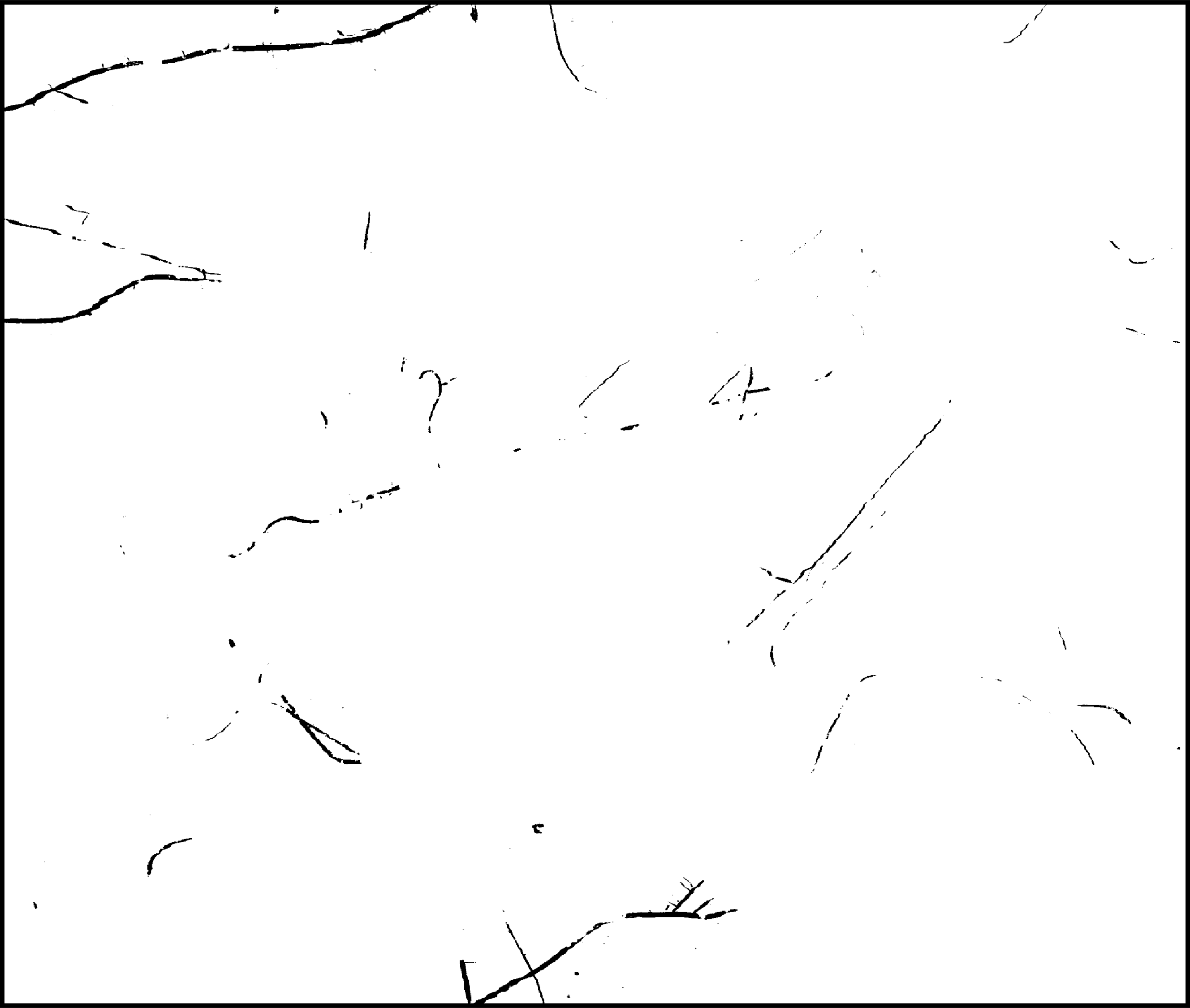}
\end{subfigure}
\begin{subfigure}{0.17\textwidth}
   \includegraphics[width=1\linewidth]{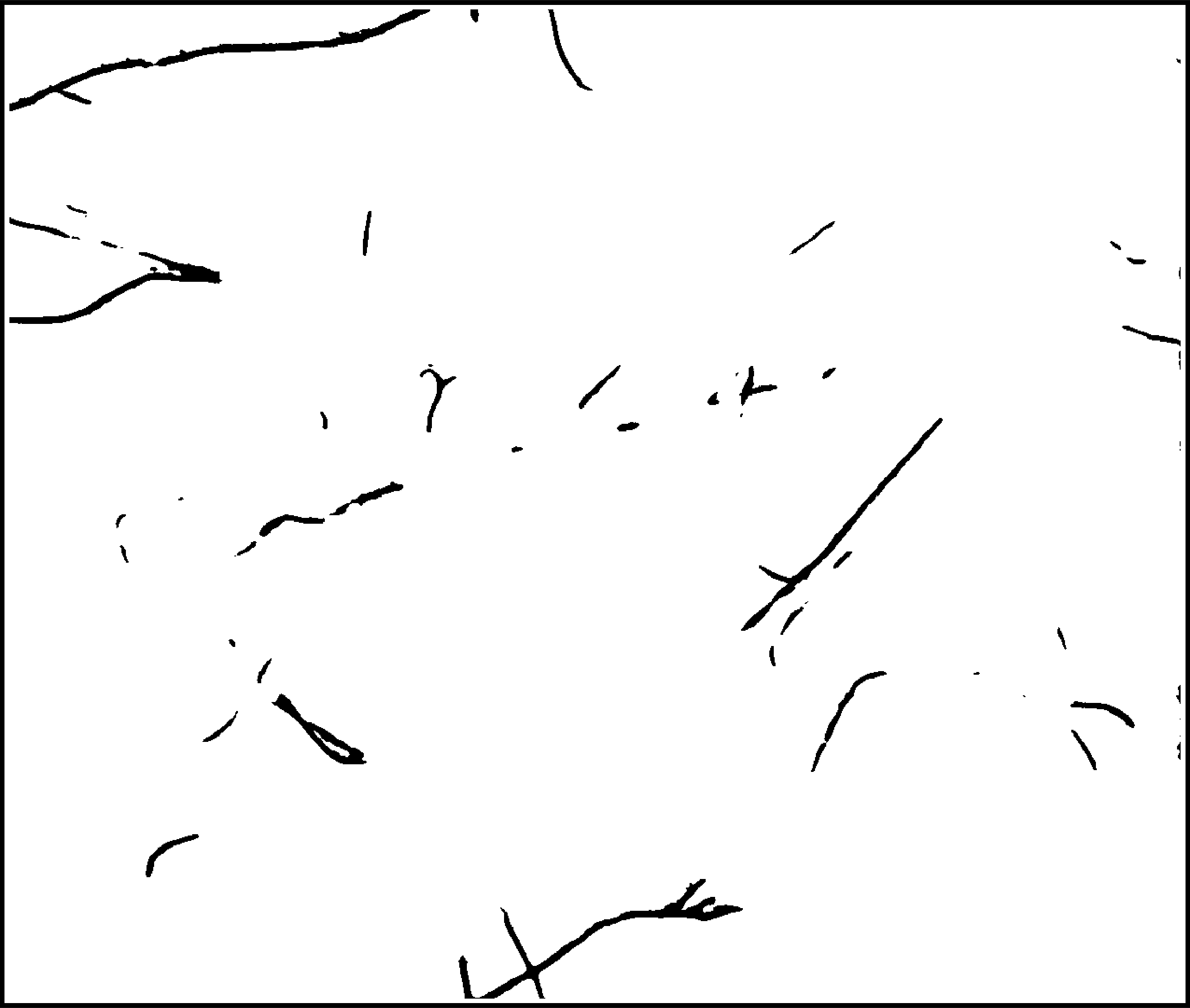}
\end{subfigure}
\\[\baselineskip]
\begin{subfigure}{0.17\textwidth}
   \includegraphics[width=1\linewidth]{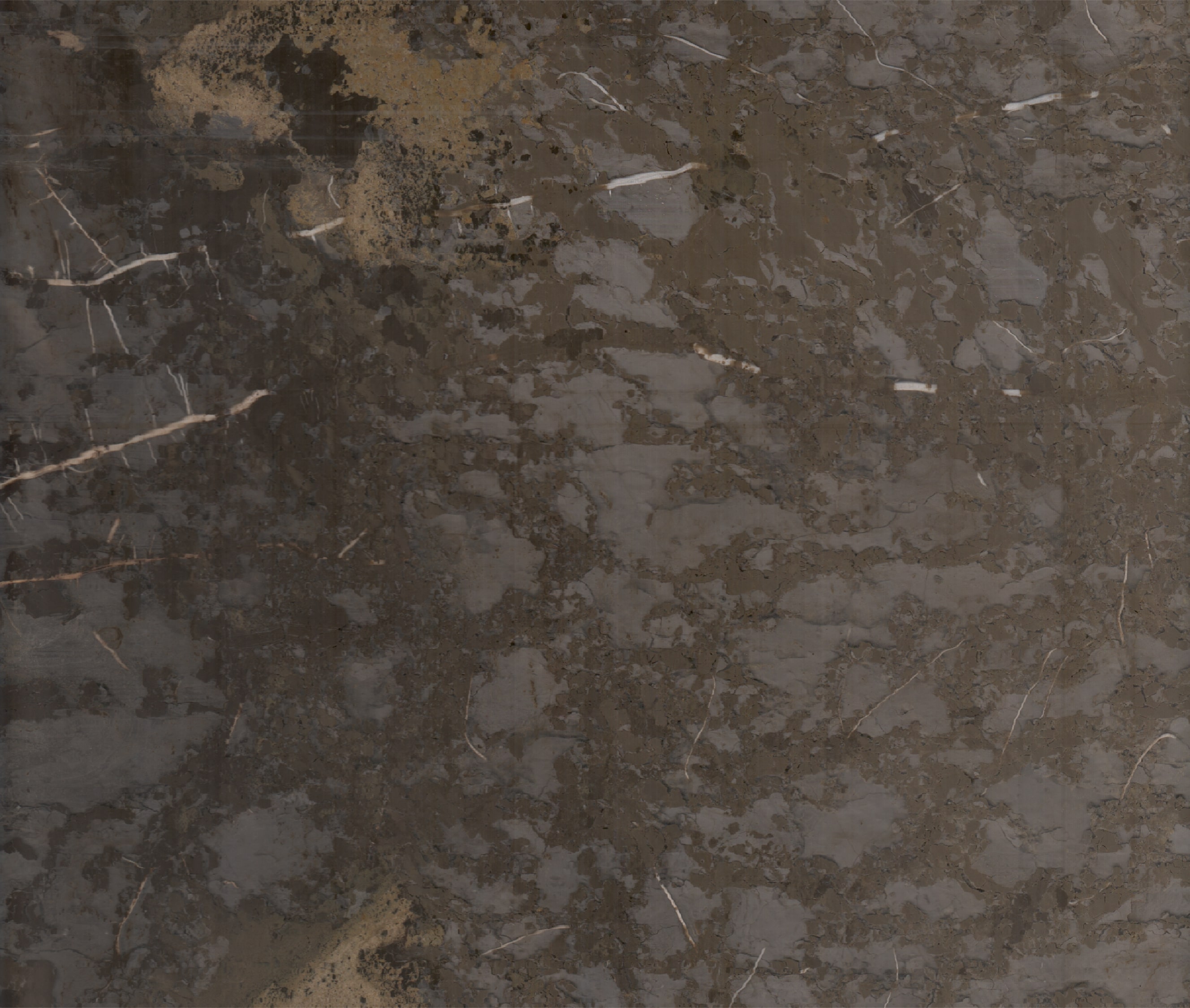}
\end{subfigure}
\begin{subfigure}{0.17\textwidth}
   \includegraphics[width=1\linewidth]{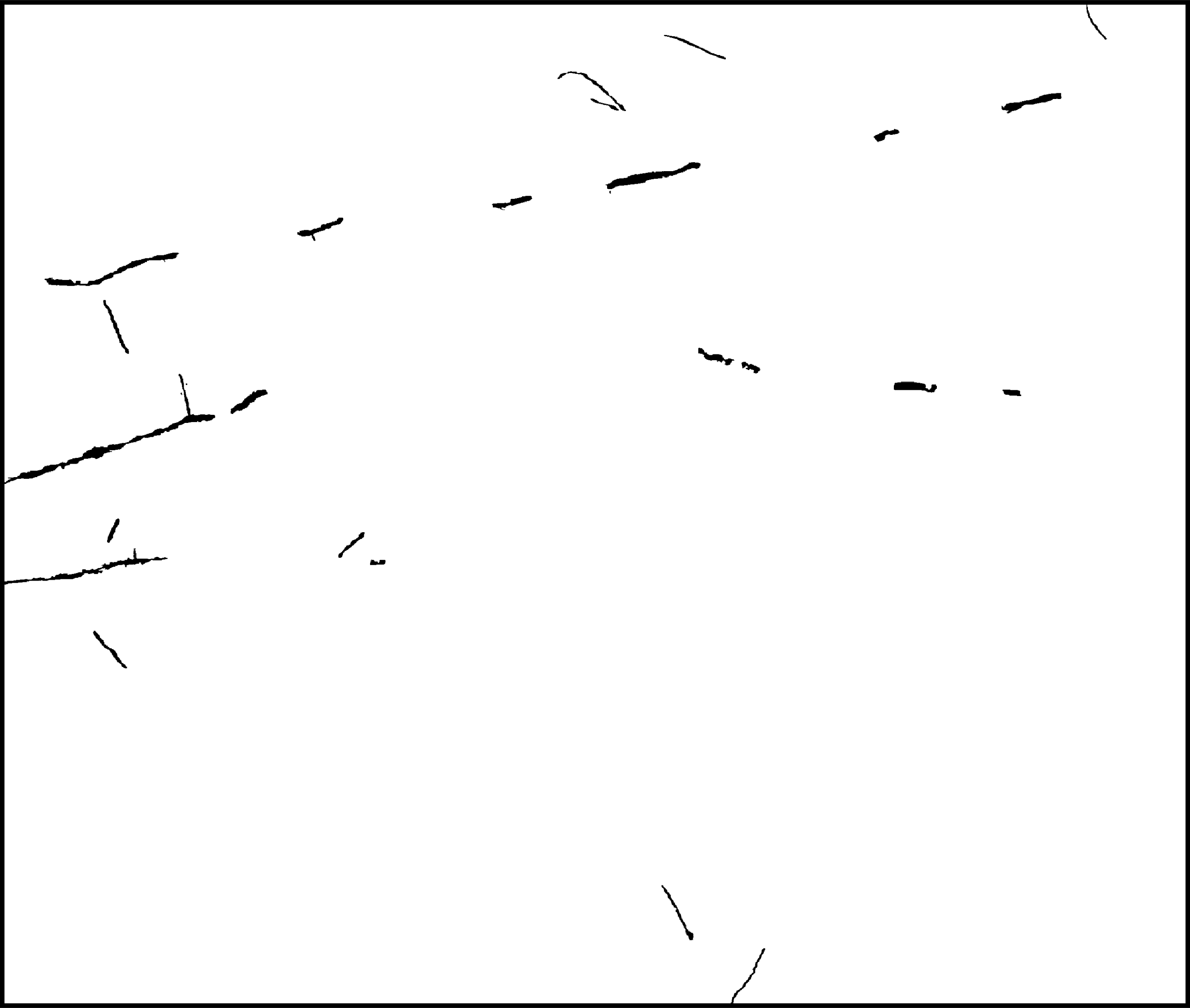}
   \end{subfigure}
\begin{subfigure}{0.17\textwidth}
   \includegraphics[width=1\linewidth]{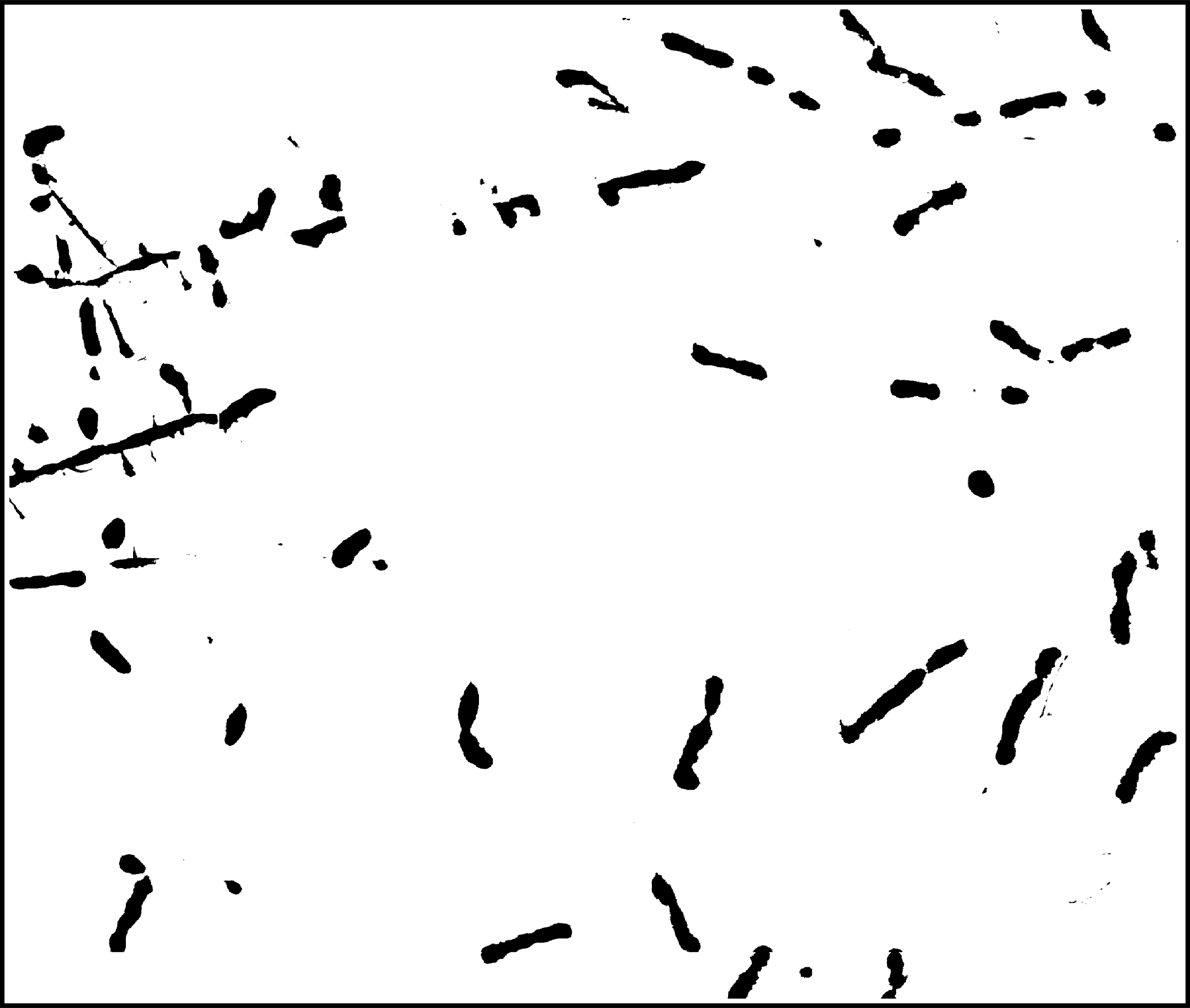}
   \end{subfigure}
\begin{subfigure}{0.17\textwidth}
   \includegraphics[width=1\linewidth]{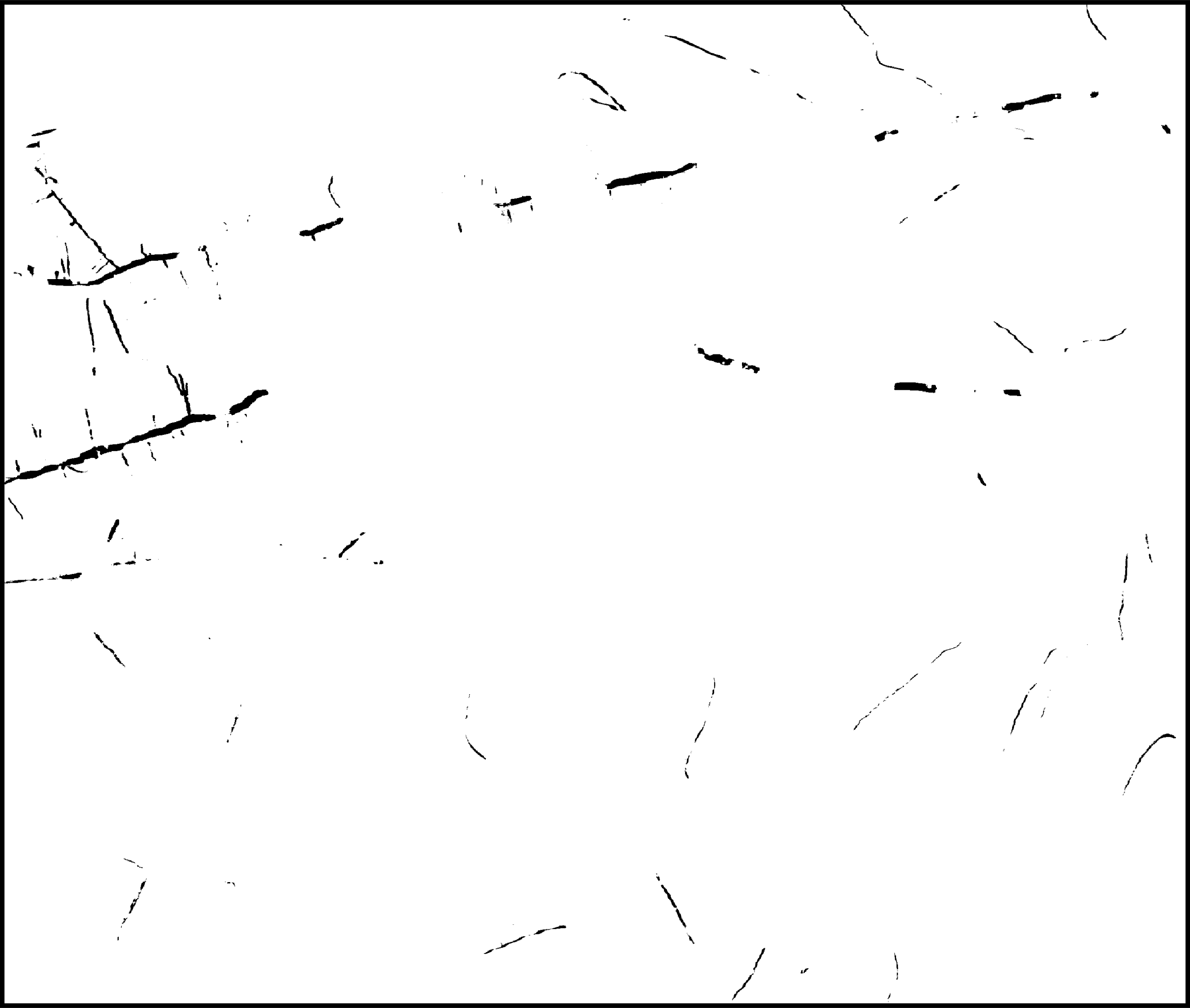}
\end{subfigure}
\begin{subfigure}{0.17\textwidth}
   \includegraphics[width=1\linewidth]{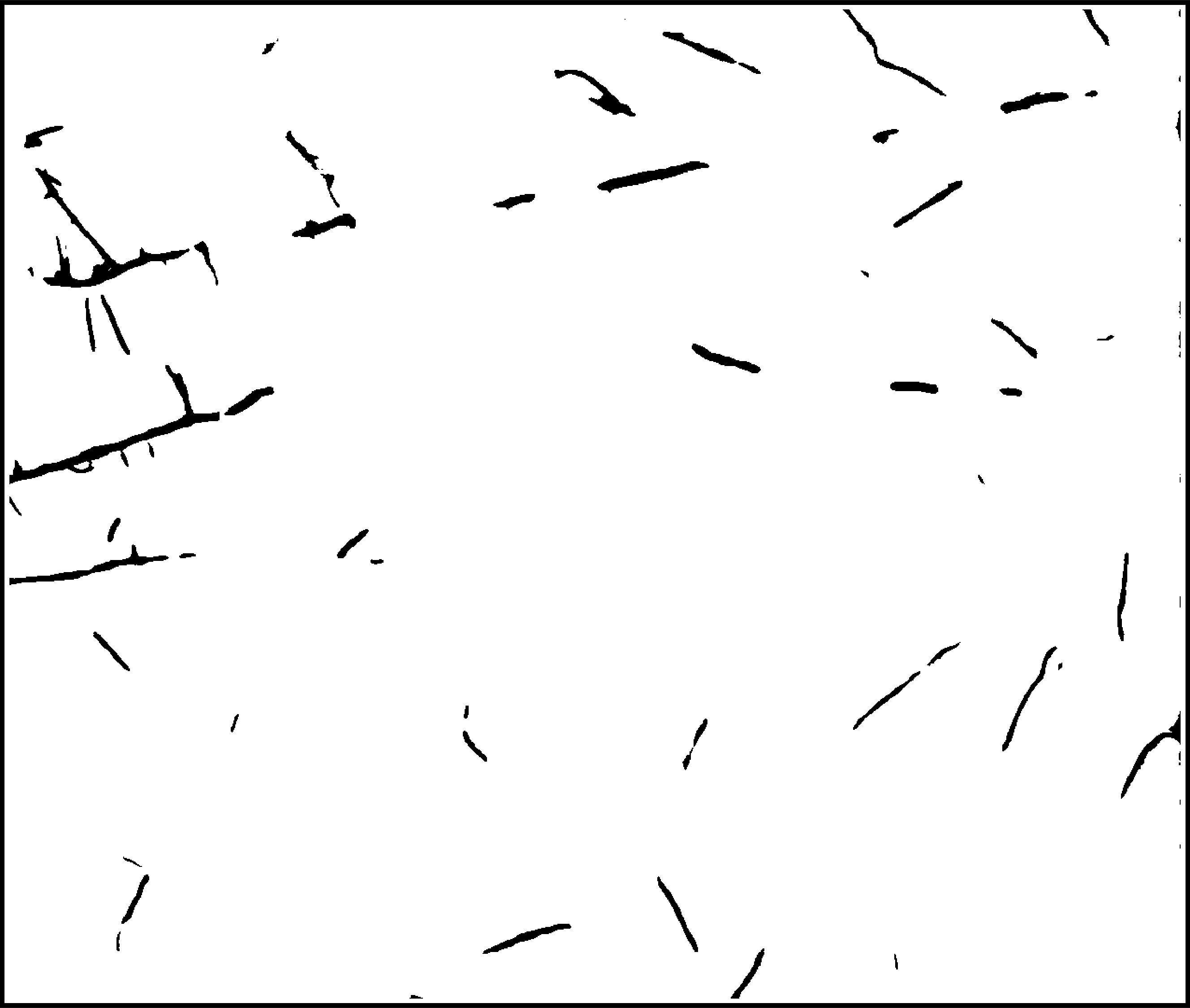}
\end{subfigure}
\\[\baselineskip]
\begin{subfigure}{0.17\textwidth}
   \includegraphics[width=1\linewidth]{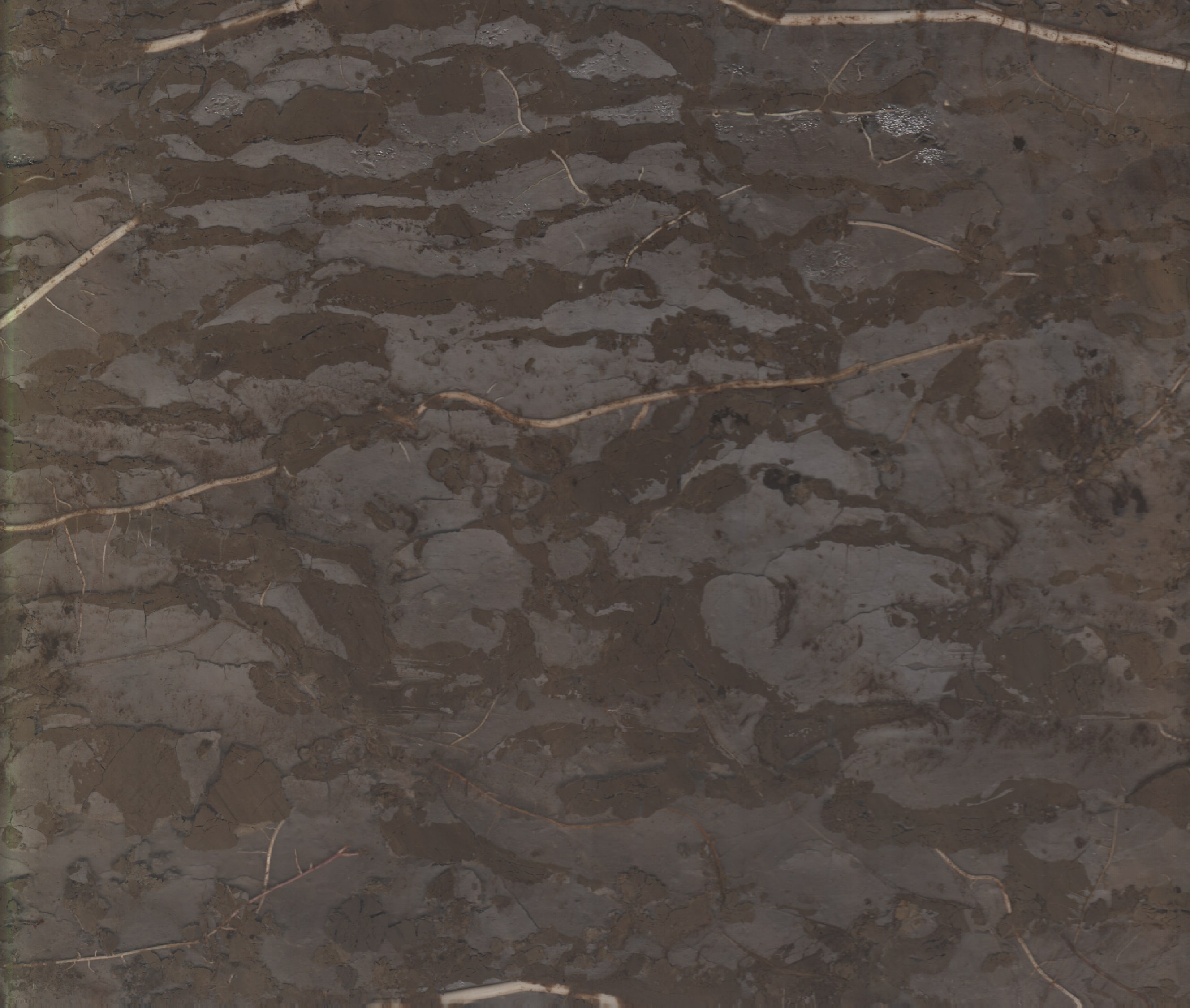}
\end{subfigure}
\begin{subfigure}{0.17\textwidth}
   \includegraphics[width=1\linewidth]{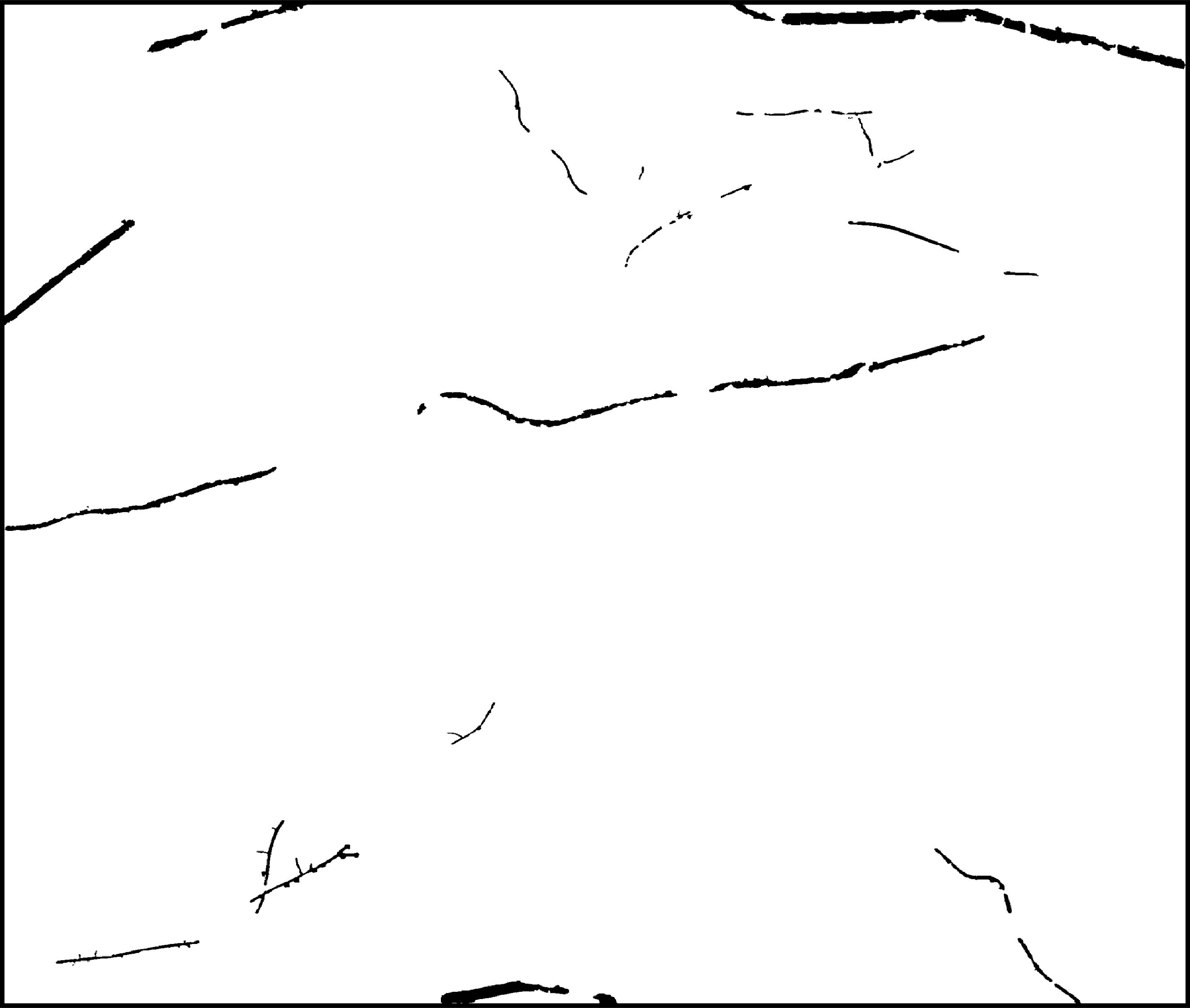}
   \end{subfigure}
\begin{subfigure}{0.17\textwidth}
   \includegraphics[width=1\linewidth]{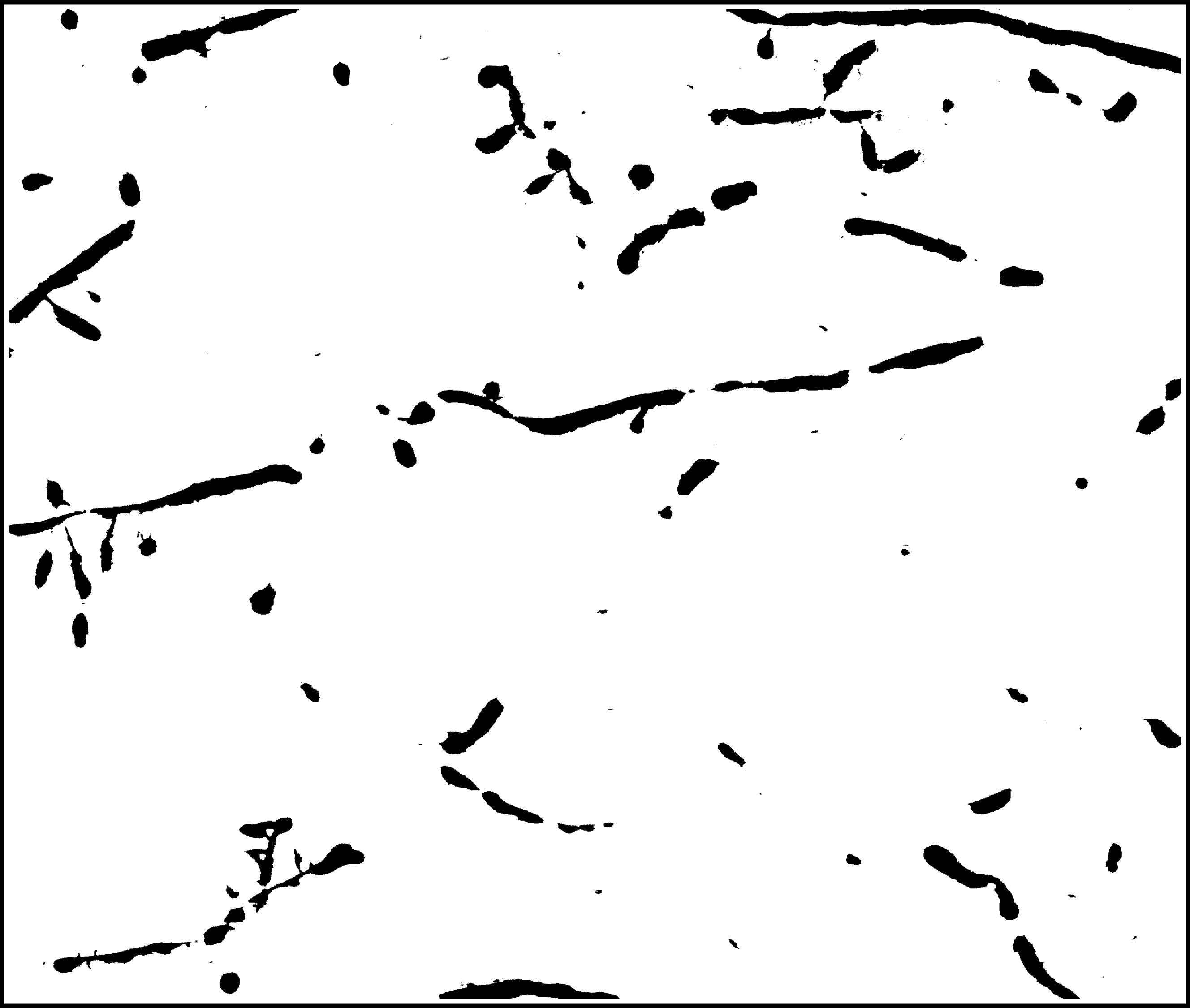}
   \end{subfigure}
\begin{subfigure}{0.17\textwidth}
   \includegraphics[width=1\linewidth]{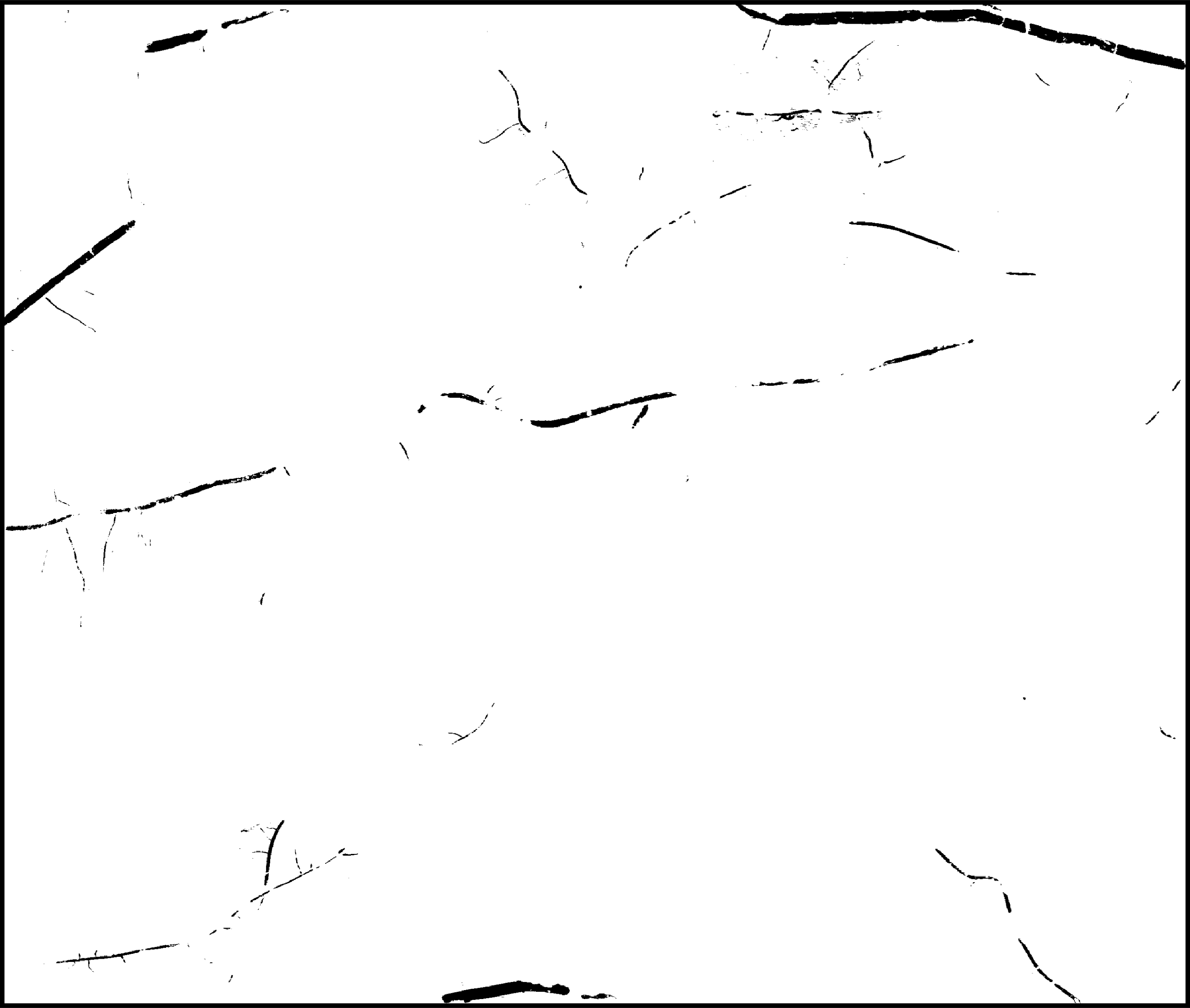}
\end{subfigure}
\begin{subfigure}{0.17\textwidth}
   \includegraphics[width=1\linewidth]{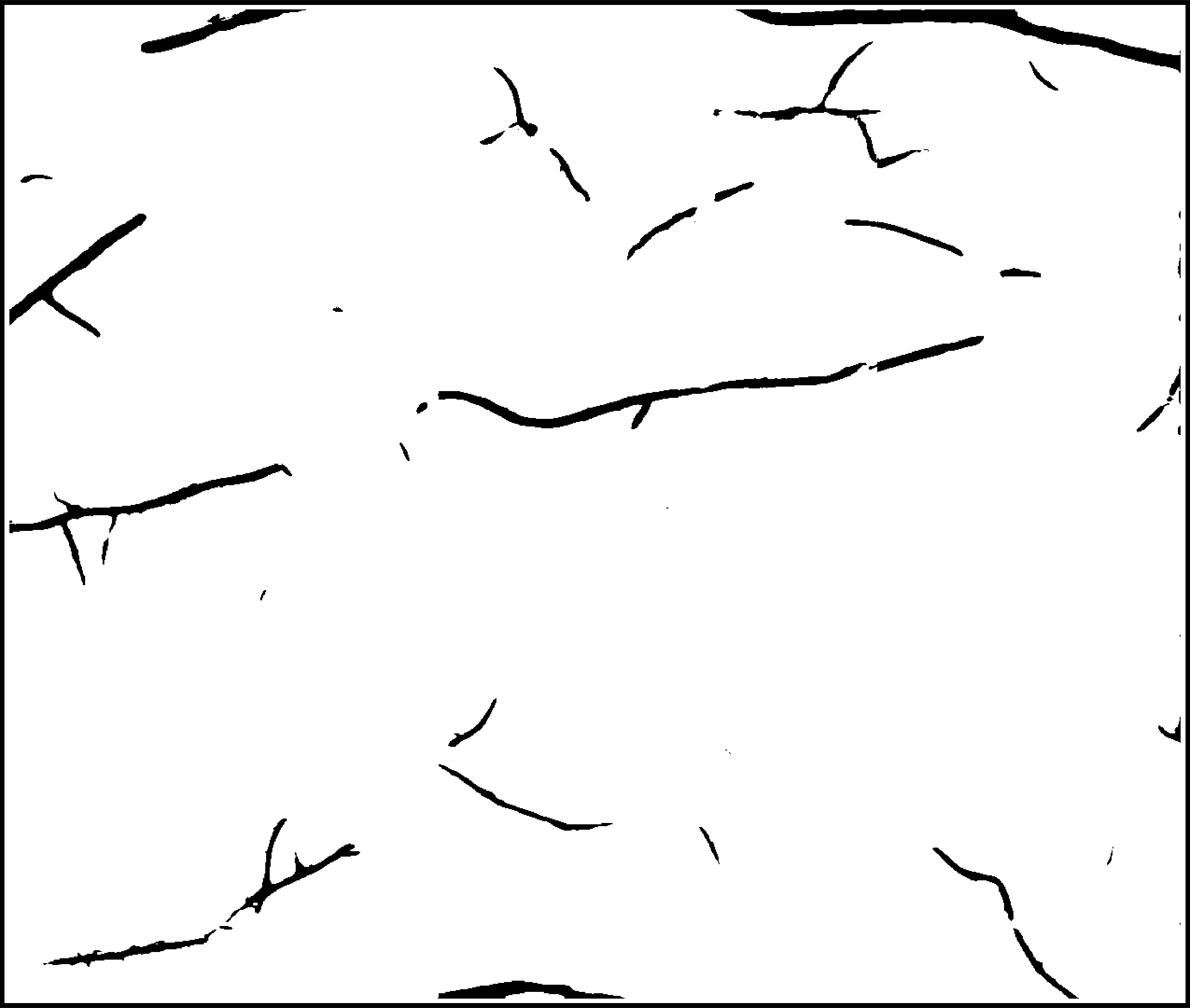}
\end{subfigure}
\\[\baselineskip]
\begin{subfigure}{0.17\textwidth}
   \includegraphics[width=1\linewidth]{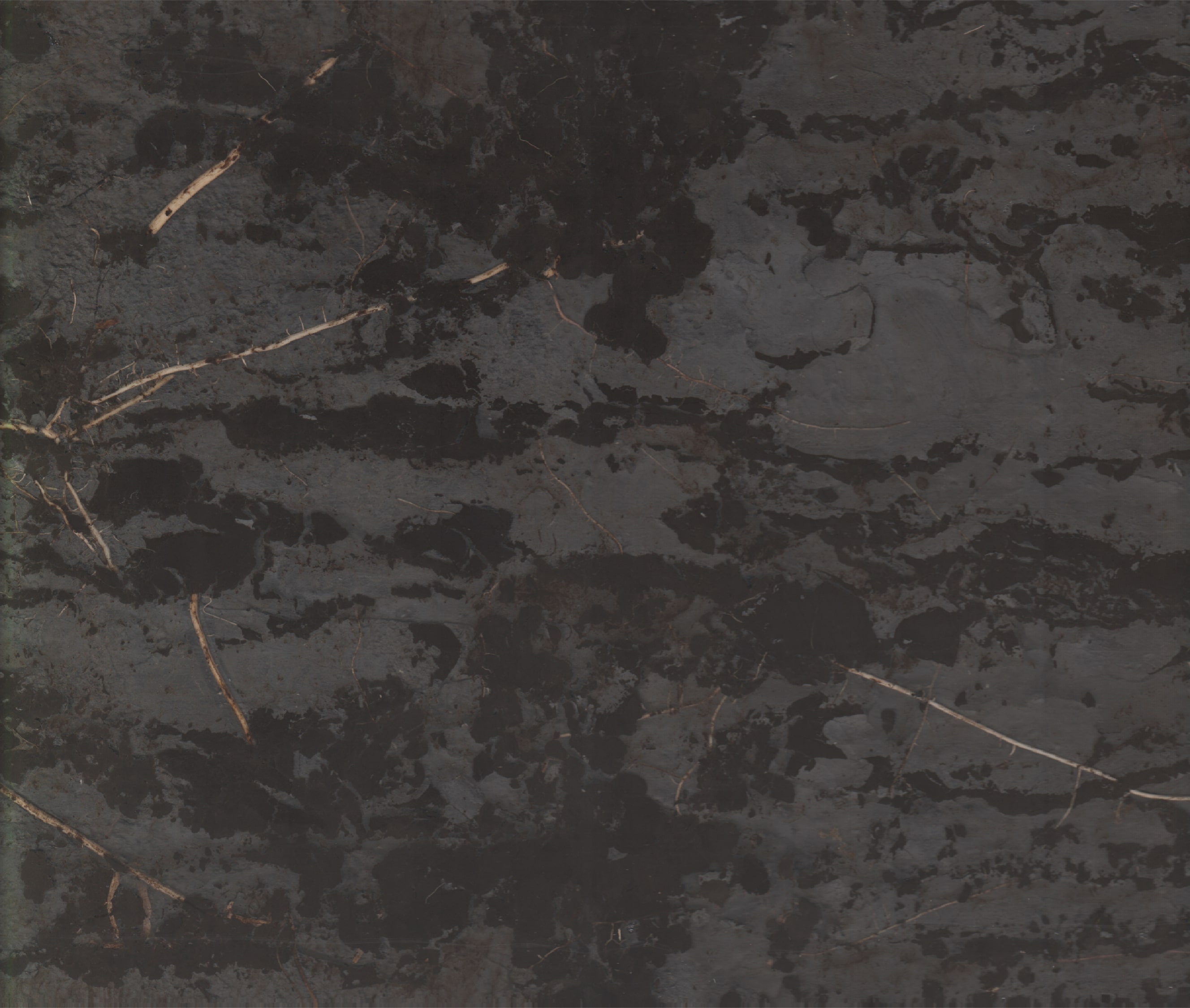}
\end{subfigure}
\begin{subfigure}{0.17\textwidth}
   \includegraphics[width=1\linewidth]{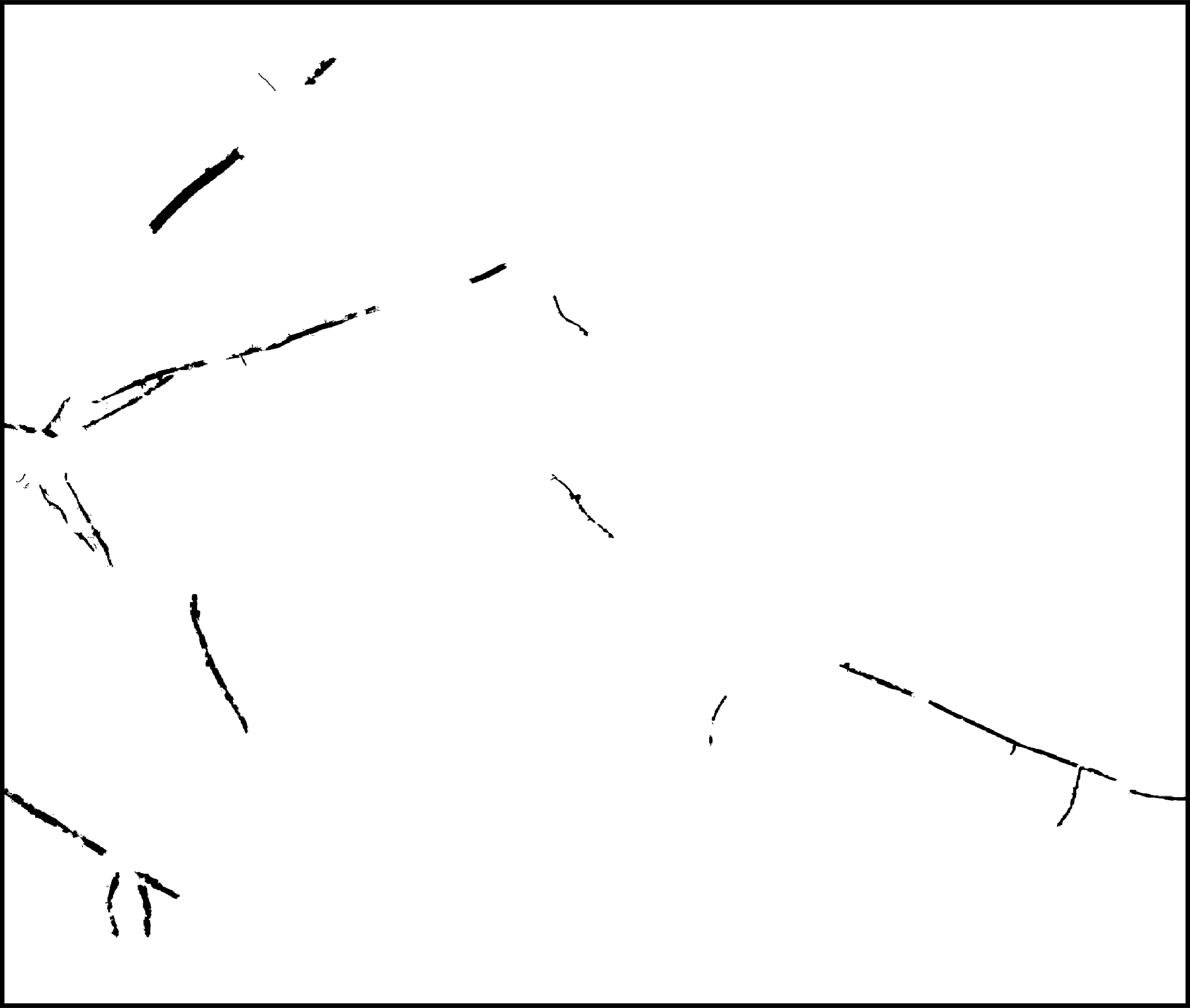}
   \end{subfigure}
\begin{subfigure}{0.17\textwidth}
   \includegraphics[width=1\linewidth]{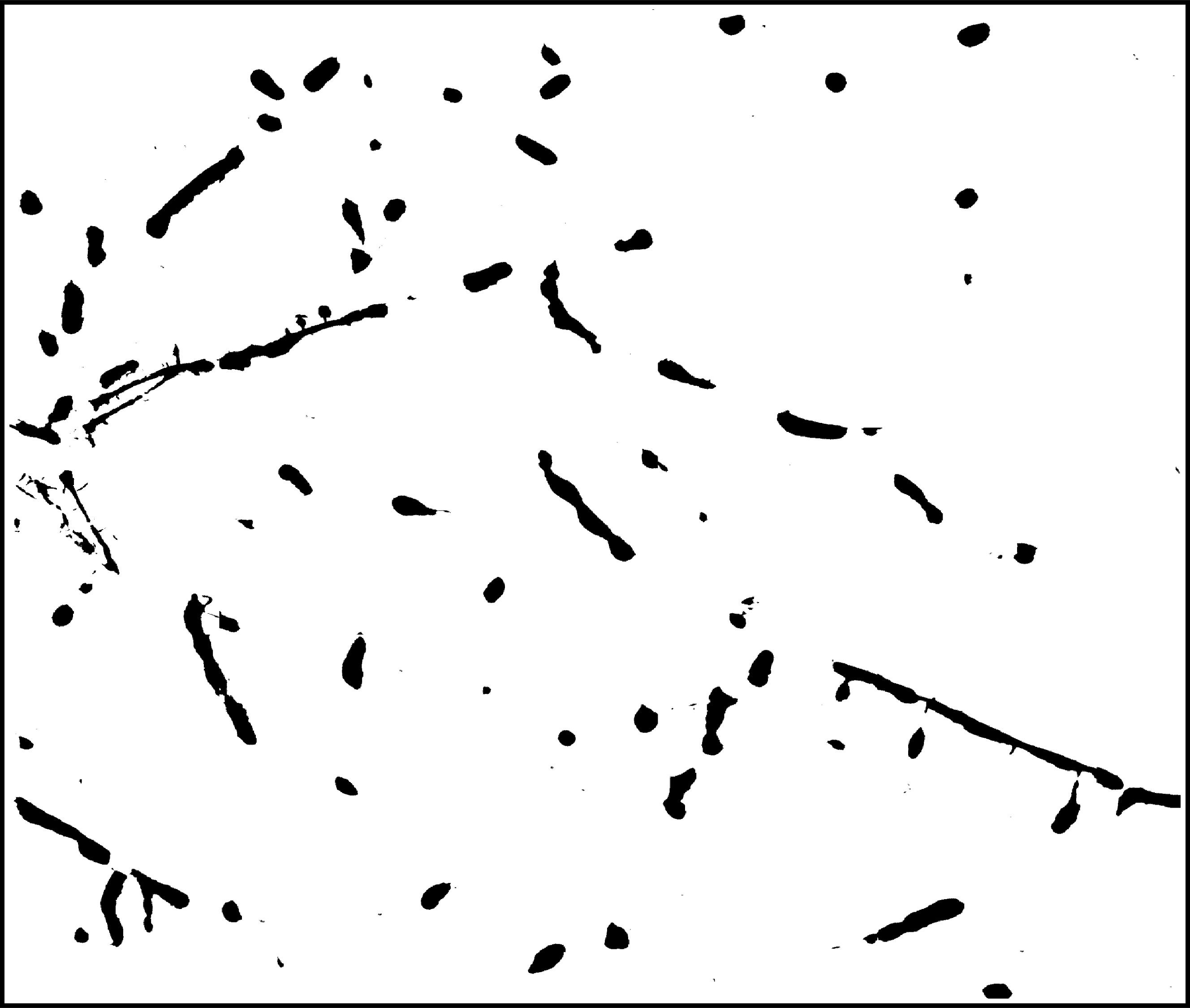}
   \end{subfigure}
\begin{subfigure}{0.17\textwidth}
   \includegraphics[width=1\linewidth]{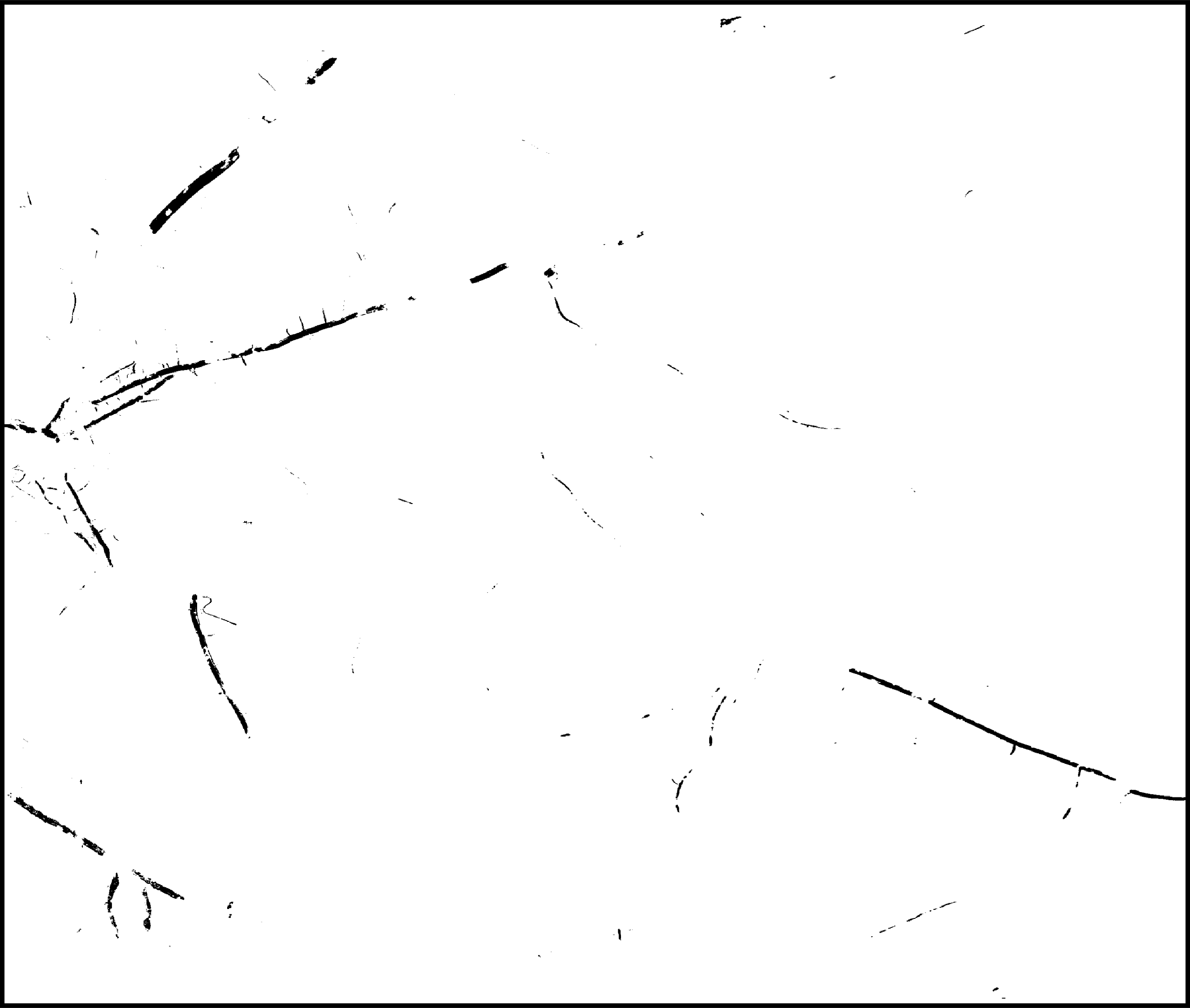}
\end{subfigure}
\begin{subfigure}{0.17\textwidth}
   \includegraphics[width=1\linewidth]{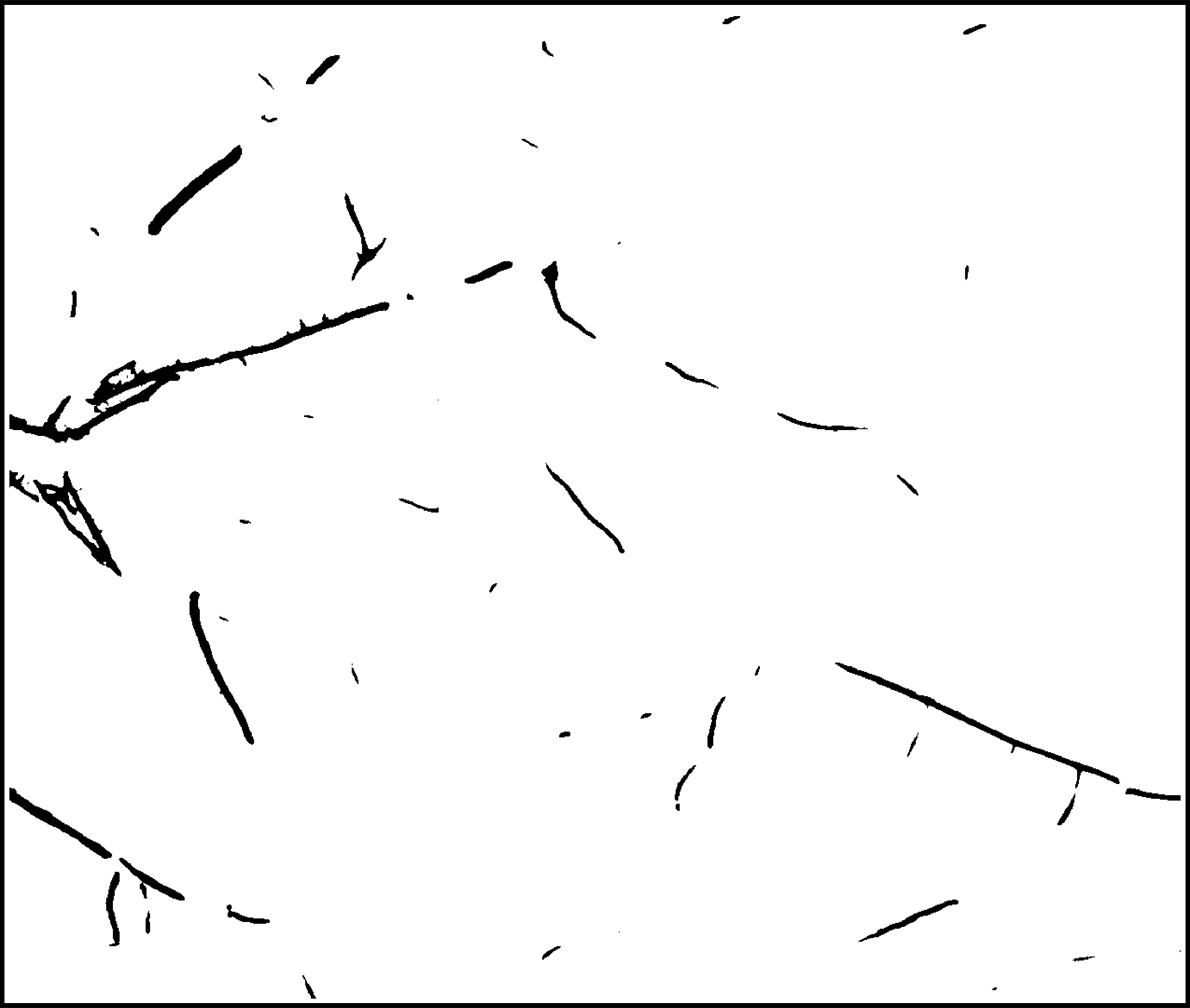}
\end{subfigure}
\\[\baselineskip]
\begin{subfigure}{0.17\textwidth}
   \includegraphics[width=1\linewidth]{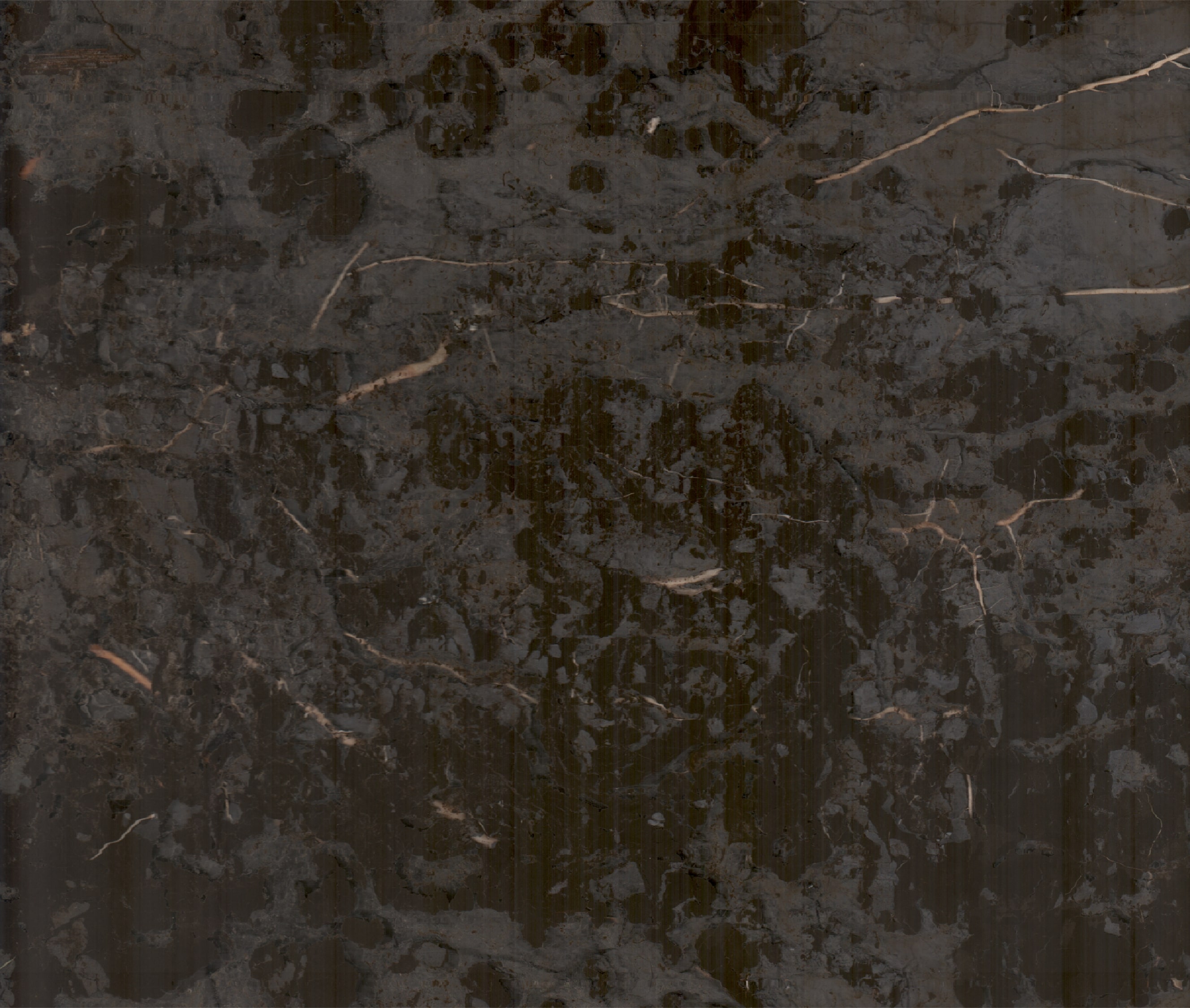}
     \caption{}
   \label{fig:4_1a}
\end{subfigure}
\begin{subfigure}{0.17\textwidth}
   \includegraphics[width=1\linewidth]{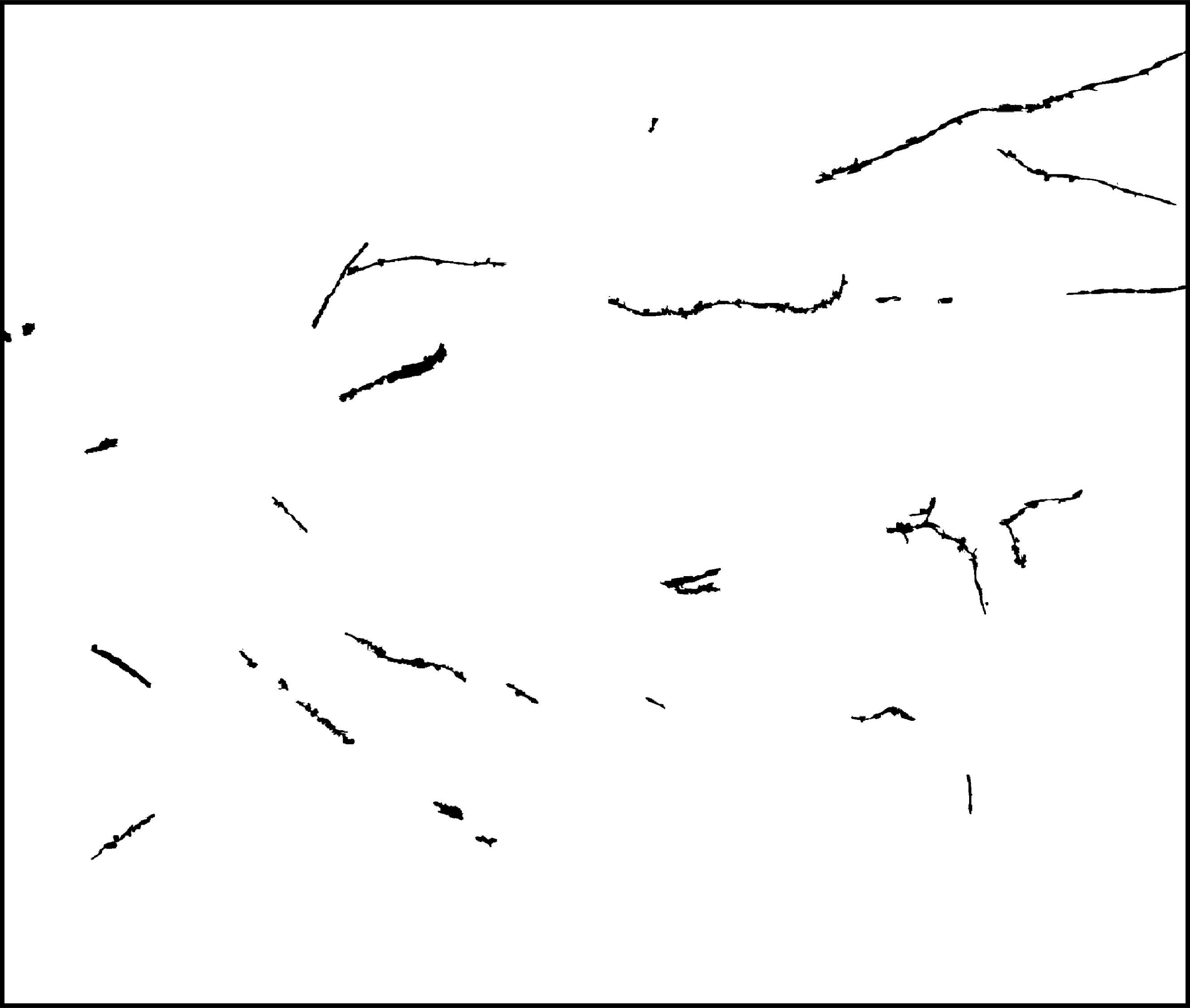}
     \caption{}
   \label{fig:4_1b}
   \end{subfigure}
\begin{subfigure}{0.17\textwidth}
   \includegraphics[width=1\linewidth]{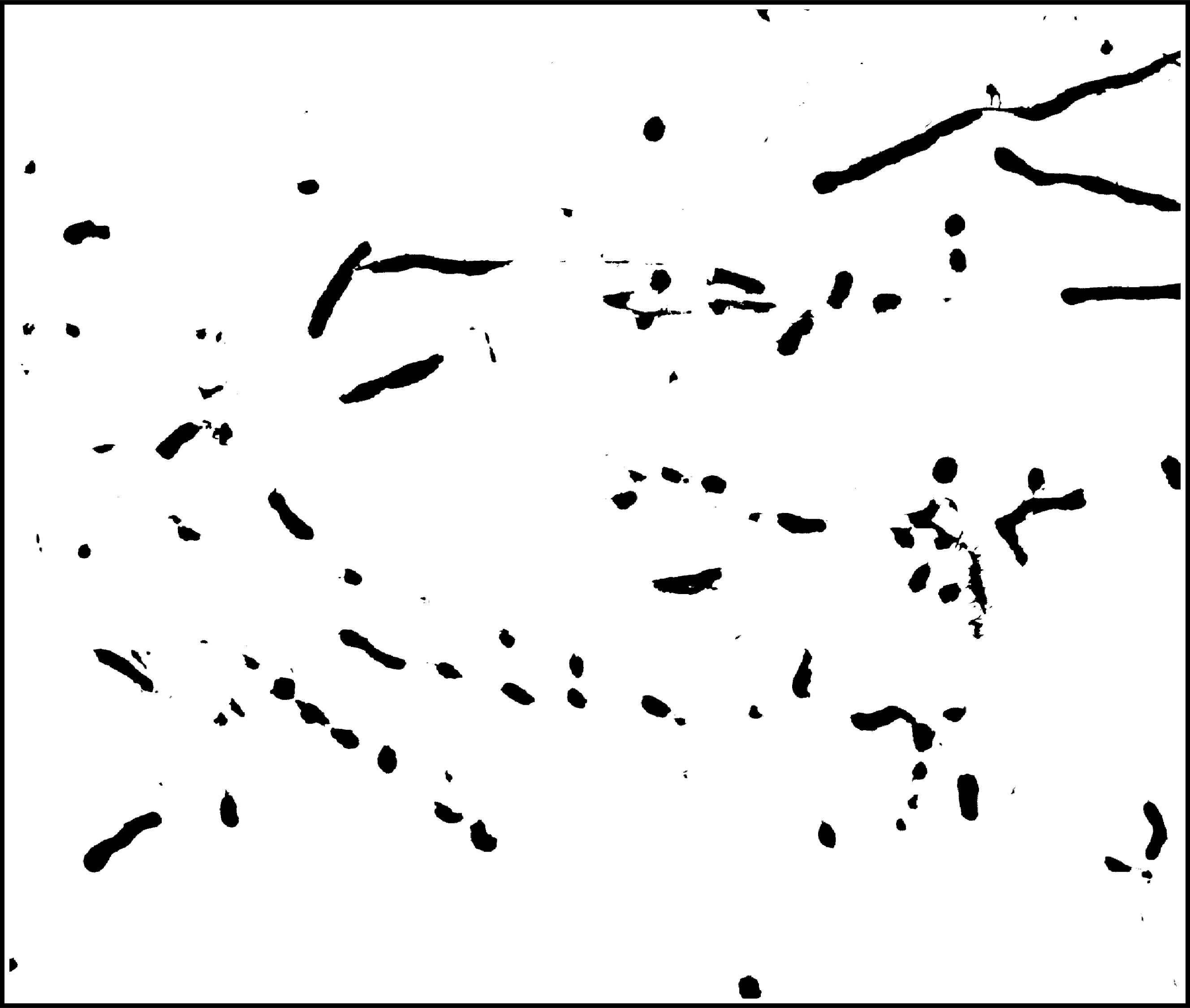}
     \caption{}
   \label{fig:4_1c}
   \end{subfigure}
\begin{subfigure}{0.17\textwidth}
   \includegraphics[width=1\linewidth]{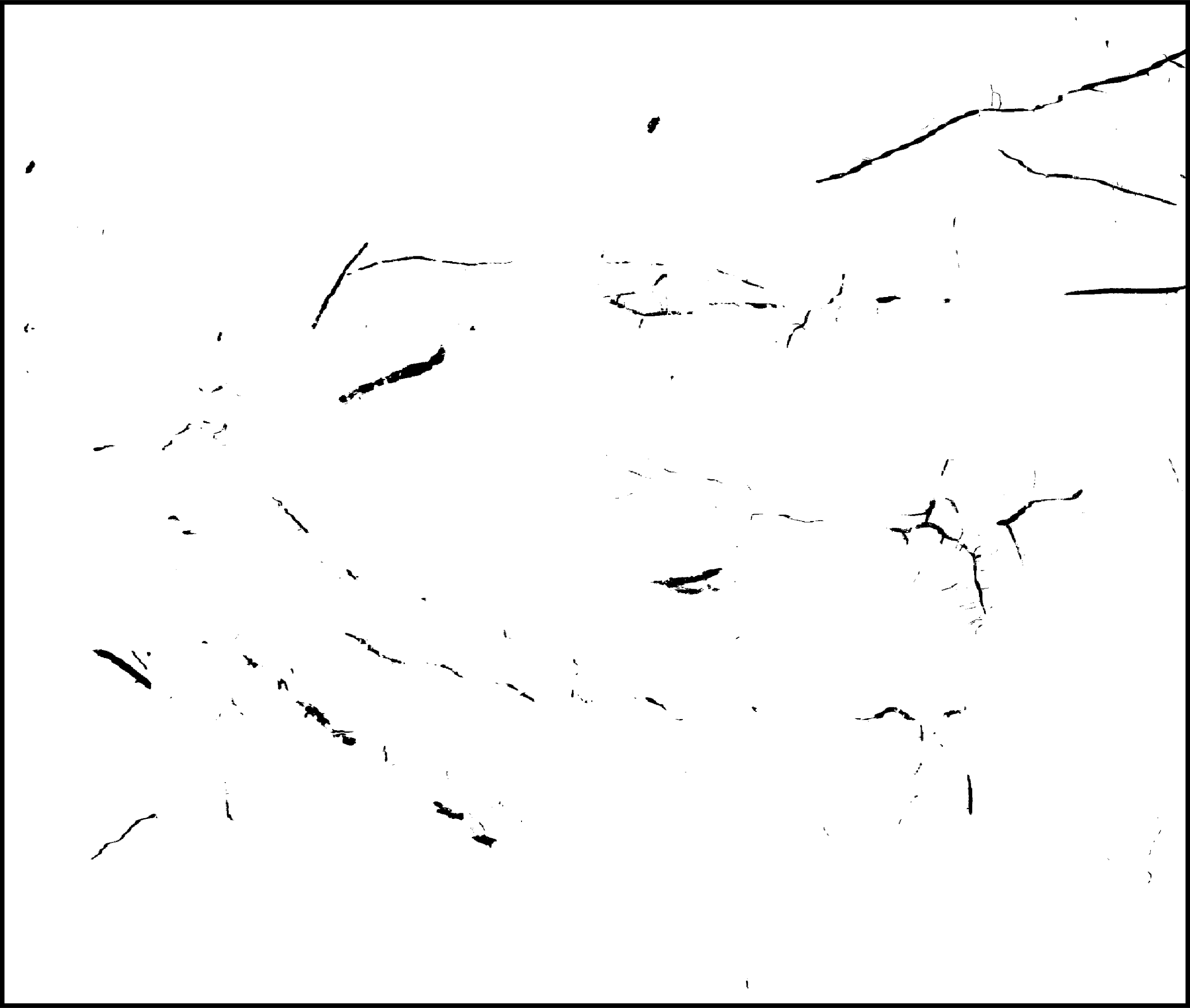}
     \caption{}
   \label{fig:4_1d}
\end{subfigure}
\begin{subfigure}{0.17\textwidth}
   \includegraphics[width=1\linewidth]{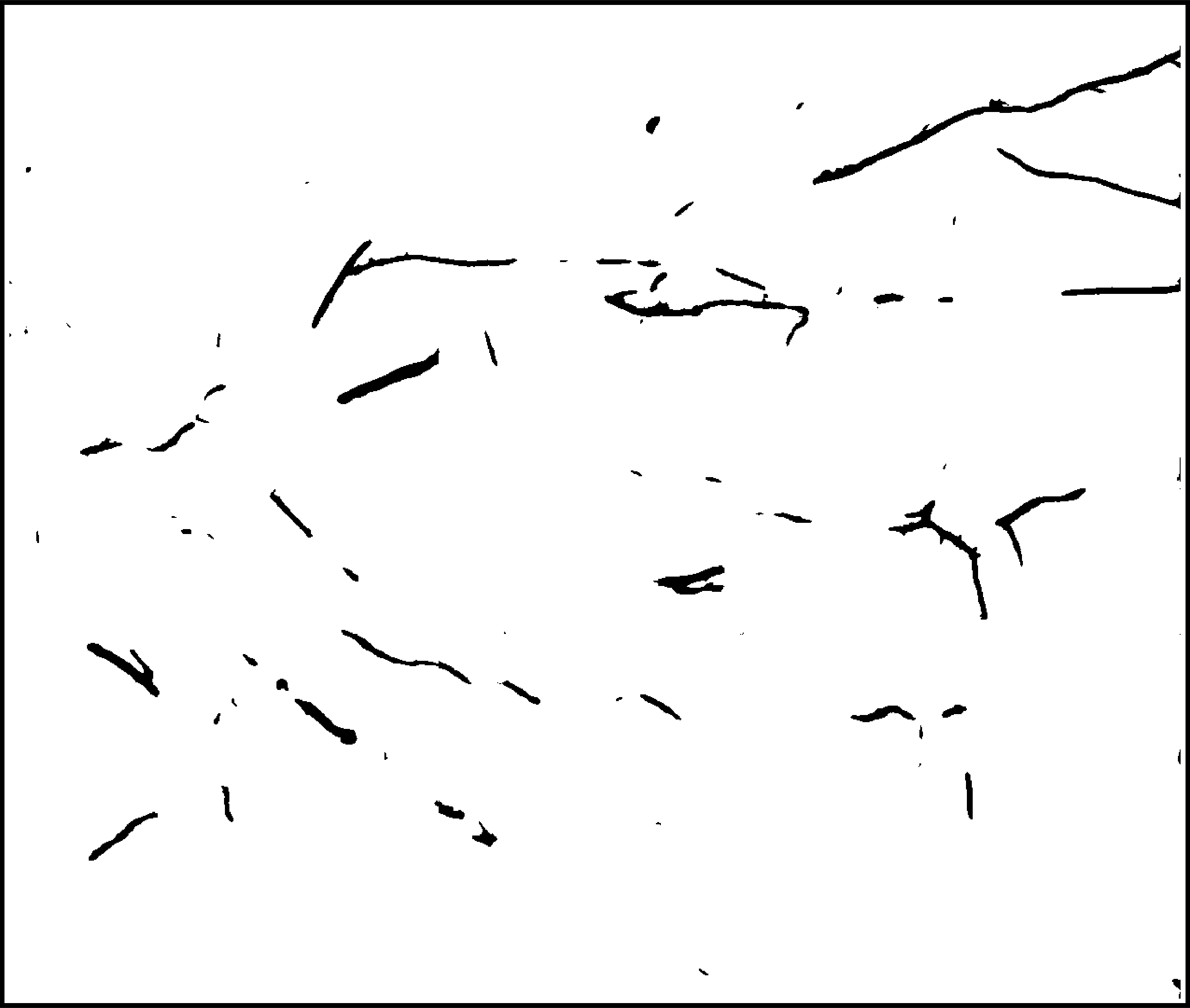}
   \caption{}
   \label{fig:4_1e}
\end{subfigure}

   \caption{Qualitative examples of root segmentation results with different method. (\subref{fig:4_1a}) Original image. (\subref{fig:4_1b}) groundtruth (GT). (\subref{fig:4_1c}) Result of argmax MIL-CAM (\subref{fig:4_1d}) Result of argmax MIL-CAM + CRF. (\subref{fig:4_1e}) Result of U-Net.}
\label{fig:4_1}
\end{center}
\end{figure*}

The segmentation results of the proposed MIL-CAM approach when taking the argmax of the softmax outputs (i.e., argmax MIL-CAM in the table) are  shown in the third column in Fig. \ref{fig:4_1c}. The long roots are a challenging problem. Although our proposed method detects most of the root pixels, it expands the boundary of some roots. This expansion results in high recall (0.859) but low precision (0.186) as shown in table \ref{table:post}. To mitigate this, we also applied a conditional random field (CRF)  \cite{krahenbuhl2011efficient} postprocessing to the segmentation results of our approach. The default parameters of the CRF were used as 0.7 for the certainty of the label, 3 for the parameter of the smoothness kernel, 80 for the spatial parameter of the appearance kernel, 13 for the color parameter of the appearance kernel and 2 inference steps were run. Segmentation results after CRF postprocessing are shown in the fourth column in in Fig. \ref{fig:4_1c}. Postprocessing improved the precision of results from 0.186 to 0.667, and the mean Intersection-Over-Union (mIoU) from 0.180 to 0.513 as shown in Table \ref{table:post}.  The only approach with that outperformed the proposed MIL-CAM with CRF postprocessing on any metric was the U-Net method outlined in \cite{xu2019overcoming}. However, this U-Net was pre-trained using a large dataset consisting of 17567 MR images with full pixel-level annotation and, thus, did not have to overcome the weak label challenge.

\section{Conclusion}
In this work, we proposed MIL-CAM for weakly supervised MR image segmentation. The proposed MIL-CAM approach outperformed a variety of comparison attention map approaches as well as a variety of MIL segmentation methods, particularly when incorporating a CRF post-processing. 

\section*{Acknowledgements}
This work was supported by the U.S. Department of Energy, Office of Science, Office of Biological and Environmental Research award number DE-SC0014156 and by the Advanced Research Projects Agency - Energy award number DE-AR0000820.

\clearpage
%
%
\bibliographystyle{splncs04}
\bibliography{egbib}
\end{document}